\documentclass[10pt,twocolumn,letterpaper]{article}

\usepackage[pagenumbers]{cvpr} %

\usepackage{booktabs}
\usepackage{tabularray}
\usepackage{multicol}
\usepackage{multirow}
\usepackage{caption}
\usepackage{graphicx}
\usepackage{listings}
\usepackage{makecell}
\usepackage{chapterbib}
\usepackage{enumitem}
\UseTblrLibrary{booktabs}

\newcommand{\Sref}[1]{Sec.~\ref{#1}}
\newcommand{\Eref}[1]{Eq.~(\ref{#1})}
\newcommand{\Fref}[1]{Fig.~\ref{#1}}
\newcommand{\Tref}[1]{Table~\ref{#1}}

\newcommand{\x}[0]{\times}
\newcommand{\fst}[1]{\textbf{#1}}
\newcommand{\snd}[1]{\underline{#1}}

\newcommand\mypara[1]{\vspace{1mm}\noindent\textbf{#1}}

\definecolor{mypink}{rgb}{0.95, 0.95, 1.0}
\definecolor{myorange}{rgb}{0.98, 0.98, 0.98}

\DeclareMathOperator*{\argmax}{arg\,max}

\definecolor{cvprblue}{rgb}{0.21,0.49,0.74}
\usepackage[pagebackref,breaklinks,colorlinks,allcolors=cvprblue]{hyperref}
\usepackage{amsmath}
\usepackage[accsupp]{axessibility}

\def\paperID{10404} %
\def\confName{CVPR}
\def\confYear{2025}

\title{CoLLM: A Large Language Model for Composed Image Retrieval}

\author{First Author\\
Institution1\\
Institution1 address\\
{\tt\small firstauthor@i1.org}
\and
Second Author\\
Institution2\\
First line of institution2 address\\
{\tt\small secondauthor@i2.org}
}

\author{
Chuong Huynh$^{1\ast}$ \quad Jinyu Yang$^{2}$ \quad Ashish Tawari$^{2}$ \quad  Mubarak Shah$^{2,3}$ \quad Son Tran$^{2}$ \\
Raffay Hamid$^{2}$ \quad Trishul Chilimbi$^{2}$ \quad Abhinav Shrivastava$^{1}$ \\
\\
$^1$~University of Maryland, College Park \\
{\tt\small \{chuonghm, abhinav\}@cs.umd.edu} \\
$^2$~Amazon \\
{\tt\small \{viyjy, atawari, sontran, raffay, trishulc\}@amazon.com} \\
$^3$~Center for Research in Computer Vision, University of Central Florida \\
{\tt\small shah@crcv.ucf.edu}
}

\author{
Chuong Huynh$^{1\ast}$ \quad Jinyu Yang$^{2\dagger}$ \quad Ashish Tawari$^{2}$ \quad  Mubarak Shah$^{2,3}$ \quad Son Tran$^{2}$ \\
Raffay Hamid$^{2}$ \quad Trishul Chilimbi$^{2}$ \quad Abhinav Shrivastava$^{1}$ \\
\\
$^1$~University of Maryland, College Park \quad $^2$~Amazon \\
$^3$~Center for Research in Computer Vision, University of Central Florida \\
{\tt\small $^1$\{chuonghm, abhinav\}@cs.umd.edu} \\
{\tt\small $^2$\{viyjy, atawari, sontran, raffay, trishulc\}@amazon.com} \quad {\tt\small $^3$shah@crcv.ucf.edu}
}

\begin{document}
\maketitle

\begin{abstract}

Composed Image Retrieval (CIR) is a complex task that aims to retrieve images based on a multimodal query. Typical training data consists of triplets containing a reference image, a textual description of desired modifications, and the target image, which are expensive and time-consuming to acquire. The scarcity of CIR datasets has led to zero-shot approaches utilizing synthetic triplets or leveraging vision-language models (VLMs) with ubiquitous web-crawled image-caption pairs. However, these methods have significant limitations: synthetic triplets suffer from limited scale, lack of diversity, and unnatural modification text, while image-caption pairs hinder joint embedding learning of the multimodal query due to the absence of triplet data. Moreover, existing approaches struggle with complex and nuanced modification texts that demand sophisticated fusion and understanding of vision and language modalities. We present CoLLM, a one-stop framework that effectively addresses these limitations. Our approach generates triplets on-the-fly from image-caption pairs, enabling supervised training without manual annotation. We leverage Large Language Models (LLMs) to generate joint embeddings of reference images and modification texts, facilitating deeper multimodal fusion. Additionally, we introduce Multi-Text CIR (MTCIR), a large-scale dataset comprising 3.4M samples, and refine existing CIR benchmarks (CIRR and Fashion-IQ) to enhance evaluation reliability. Experimental results demonstrate that CoLLM achieves state-of-the-art performance across multiple CIR benchmarks and settings. MTCIR yields competitive results, with up to 15\% performance improvement. Our refined benchmarks provide more reliable evaluation metrics for CIR models, contributing to the advancement of this important field. Project page is at \url{collm-cvpr25.github.io}.
\vspace{-1em}
\end{abstract}
\renewcommand{\thefootnote}{$\ast$}
\footnotetext[0]{This work was done during internship at Amazon. $\dagger$~Project lead.}
\renewcommand{\thefootnote}{\arabic{footnote}}

\section{Introduction}
\label{sec:intro}

\begin{figure}[t]
    \centering
    \vspace{-0.5em}
    \begin{subfigure}[b]{\linewidth}
        \centering
        \includegraphics[width=\columnwidth]{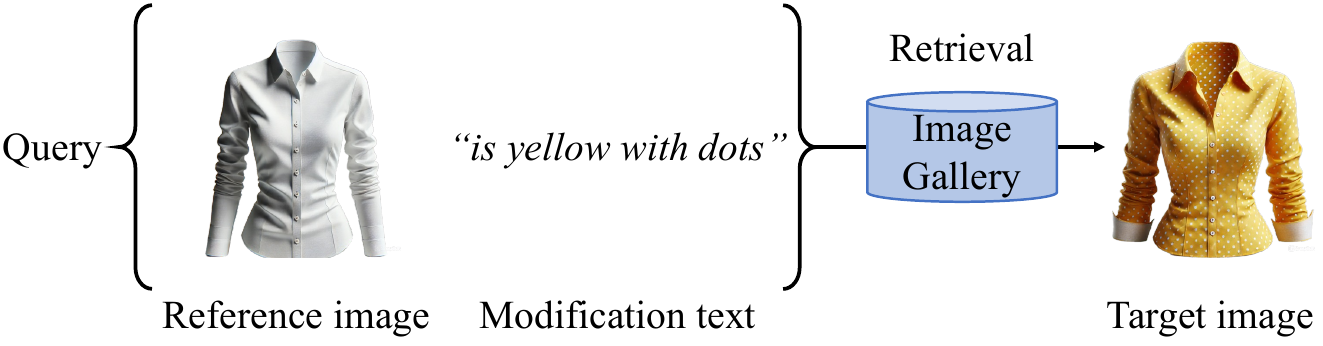}
        \caption{Composed Image Retrieval Example}
        \label{fig:teaser_example}
        \vspace{0.5em}
    \end{subfigure}
    \begin{subfigure}[b]{\linewidth}
        \includegraphics[width=\columnwidth]{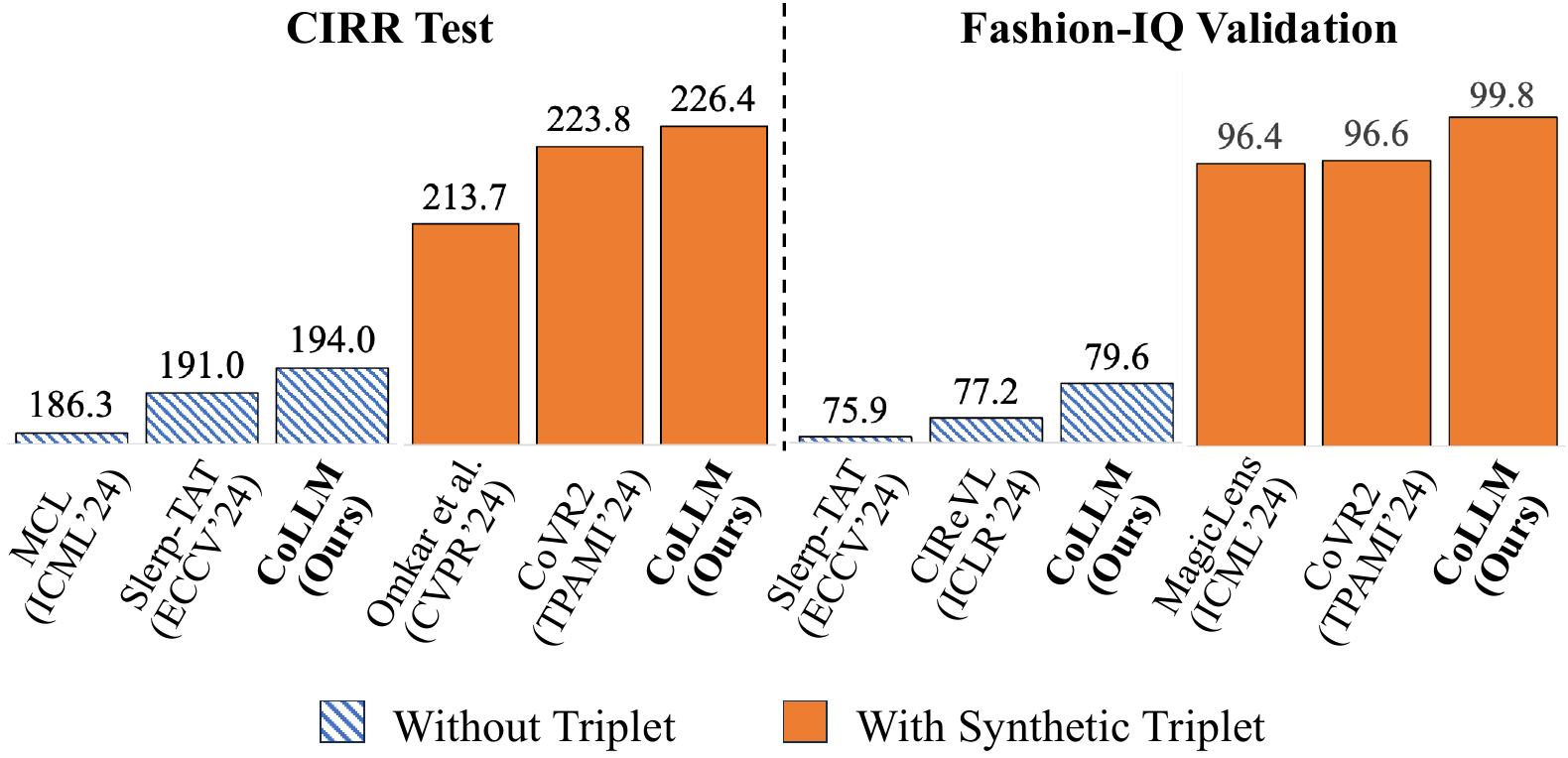}
        \caption{Recall Sum of state-of-the-art methods on CIR benchmarks}
        \label{fig:teaser_recall}
    \end{subfigure}
    \caption{
    (a) An example of CIR. (b) Recall Sum at \{1,10,50\} for CIRR and \{10, 50\} for Fashion-IQ between CoLLM and state-of-the-art (SoTA) models under zero-shot settings. We evaluate two training scenarios: (i) without triplet data and (ii) with synthetic triplet data.
    }
    \label{fig:teaser}
    \vspace{-10pt}
\end{figure}

Composed Image Retrieval (CIR) enhances traditional image retrieval~\cite{cora,yang2022vision} by combining text and image queries, offering greater flexibility in search systems, with applications in e-commerce, fashion, and design ~\cite{deldjoo2023review,goenka2022fashionvlp,simpson,yuqian2024machine,zhu2024bringing}. 
As illustrated in ~\Fref{fig:teaser_example}, CIR retrieves similar items, such as shirts, where the reference image provides a visual basis, and the modification text specifies desired modifications. 
By expanding the capabilities of conventional image or text-based searches, CIR allows for more nuanced and specific queries. 
However, this advanced approach also presents significant challenges compared to traditional image retrieval.

Supervised CIR approaches face challenges in data acquisition. They require high-quality CIR triplets (reference image, target image, and modification text), collected through a labor-intensive and expensive process. 
Consequently, existing CIR triplet datasets are limited in scale and domain coverage, restricting model generalizability.
To overcome these limitations, recent CIR methods~\cite{caselasco,compodiff,magiclens,ventura24covr2,vdg} have adopted zero-shot approaches~\cite{liu2023zero}. 
These can be broadly categorized into two strategies: (i) leveraging vision-language models~\cite{clip,blip,sirnam2024x} (VLMs) for composed query embedding generation, and (ii) generating synthetic triplets for supervised training.
VLM-based approaches utilize pre-trained models that already align vision and language features in latent space. These methods rely on large-scale image-caption pairs and follow two main directions, i.e., textual inversion~\cite{contexti2w,lincir,palavra} and direct interpolation~\cite{slerptat}.
Synthetic triplet generation approaches~\cite{magiclens,caselasco,vdg} employ Large Language Models (LLMs)~\cite{touvron2023llama,zhang2022opt,chowdhery2023palm} to generate modification text, with CompoDiff~\cite{compodiff} further utilizing diffusion models~\cite{rombach2022high,brooks2023instructpix2pix} to create synthetic reference-target image pairs. Despite the recent advances in CIR, several critical challenges persist:
\begin{itemize}
\item \textbf{[Data] Limitation-1:} VLM-based methods may hinder efficient and effective composed query embedding learning by relying solely on image-caption pairs. 
\item \textbf{[Data] Limitation-2:} Existing synthetic triplet datasets often suffer from a lack of diversity and unnatural modification text. Moreover, these datasets are either very small or closed-source, impeding research progress in the field.

\item \textbf{[Model] Limitation-3:} Current methods for composed query embeddings mainly use shallow transformer models or linear interpolation. While these methods are computationally efficient, they lack the ability to capture the full complexity of the composed understanding tasks.
\item \textbf{[Evaluation] Limitation-4:} Existing CIR benchmarks are often compromised by noise, particularly in the form of ambiguous samples. Ambiguity arises when multiple target images can correctly match a single query, but only one is labeled as the ground truth, ignoring valid matches. Although CIRCO~\cite{searle} attempts to address this issue, its efforts are limited in scale. This ambiguity in benchmarks hinders meaningful model evaluation and comparison.
\end{itemize}
We propose CoLLM, an LLM-based CIR approach to address the aforementioned limitations. CoLLM tackles \textbf{Limitation-1} by dynamically synthesizing triplets from image-caption pairs, introducing two key components: a reference image embedding synthesis and a modification text synthesis module. The former employs Spherical Linear Interpolation~\cite{slerp} (Slerp) to generate an intermediate embedding between a given image and its nearest neighbor within the training batch, serving as a synthesized reference image embedding. The latter utilizes a pre-defined text interpolation template to generate modification text based on the current caption and that of the nearest image neighbor. This strategy effectively leverages the vast amount of readily available image-caption pairs on the Internet, enabling a model's training in a supervised CIR manner. By doing so, CoLLM overcomes the scarcity of labeled triplet data and paves the way for more robust and scalable CIR models.

To address \textbf{Limitation-2}, we introduce Multi-Text CIR (MTCIR), a synthetic dataset of 3.4M image pairs with 17.7M modification texts. MTCIR focuses on two often-overlooked aspects: image diversity and naturalistic modification texts. We curated images from diverse sources to ensure variety. For modification text generation, we employ a two-stage approach using Multi-modal LLM (MLLM)~\cite{liu2023llava,liu2024llavanext} for detailed captioning and LLM for describing inter-caption differences. Uniquely, MTCIR provides multiple short modification texts for each image pair, covering various attributes. 
This better reflects human query formulation, offering a more realistic, comprehensive training foundation for CIR models.

To overcome \textbf{Limitation-3}, CoLLM harnesses the power of LLMs for composed query understanding. It is motivated by the extensive world knowledge embedded in pre-trained LLMs. With their deep semantic understanding, we posit that LLMs are superior to shallow transformers and simple embedding interpolation techniques in comprehending the intricate relationships between reference images and modification texts. By leveraging LLMs, CoLLM aims to capture nuanced semantic connections, enhancing the quality and relevance of composed image retrieval results. 

To address \textbf{Limitation-4}, we refine two popular CIR benchmarks: CIRR~\cite{cirr} and Fashion-IQ~\cite{fashioniq}. We use MLLMs to evaluate sample ambiguity in each benchmark and regenerate clear modification text for ambiguous ones. Our pipeline incorporates multiple validation steps to guarantee the enhanced quality of the refined samples.

Our main contributions can be summarized as: (i) We propose a method to synthesize CIR triplets on-the-fly from image-caption pairs, which can even outperform models trained on CIR triplets, eliminating the need for costly CIR datasets. (ii) We collect and will release MTCIR, a new CIR triplet dataset covering 3.4 million image pairs with 17.7 million modification texts. To our knowledge, MTCIR is the largest open-source synthetic CIR dataset. (iii) We introduce an approach to leverage LLMs for composed query understanding, utilizing their instruction-following and embedding generation capabilities. (iv) We refine CIRR and Fashion-IQ to provide more robust evaluation benchmarks for the CIR community. Extensive experiments on popular CIR benchmarks and settings demonstrate the effectiveness of our model innovations and new datasets (~\Fref{fig:teaser_recall}).

\vspace{-5pt}
\section{Related Works}
\label{sec:related_works}
\mypara{Composed Image Retrieval.} Composed Image Retrieval (CIR) has garnered significant attention due to its flexibility in search systems~\cite{cirr,psomas2024composed,saha2018towards}. Zero-shot CIR methods have been extensively explored, with textual inversion~\cite{pic2word,searle,contexti2w,palavra,lincir} emerging as a prominent technique. This approach maps image encoder outputs to text encoder inputs, creating pseudo-word tokens. SlerpTAT~\cite{slerptat} emphasizes the text encoder's importance by re-aligning visual and textual embeddings. While most zero-shot CIR models utilize image-caption pairs for training, our work introduces an innovative method to synthesize triplets from these pairs during training. This method enables supervised CIR style training, enhancing the model's understanding of composed queries without relying on real CIR datasets.

Recent works enhance enhancing models' language understanding using Large Language Models (LLMs)~\cite{mcl,cirevl}, primarily leveraging their text generation abilities. 
MCL~\cite{mcl} composes image-text query in the LLM input space and aligns joint embedding with the target image caption.
However, this may introduce noise due to limited visual information in the target image captions. CIReVL~\cite{cirevl} employs LLMs to generate captions based on the reference image and modification text, subsequently retrieving the target image using a text-to-image retrieval approach. Our model also utilizes LLMs but differs by directly producing a composed query embedding for target image retrieval, potentially reducing intermediate steps and associated errors.

\mypara{CIR with Synthetic Datasets.} The scarcity of supervised CIR datasets has prompted recent studies to leverage generative models for synthetic triplet creation. CompoDiff~\cite{compodiff} employs image editing pipelines~\cite{rombach2022high,brooks2023instructpix2pix} to generate target images, though their nature limits performance. Other approaches~\cite{magiclens,caselasco,vdg,webcovr,ventura24covr2} utilize LLMs to generate modification text for real image pairs. Our work introduces a novel two-stage approach, combining MLLMs~\cite{liu2023llava,liu2024llavanext} for detailed captioning and LLMs~\cite{achiam2023gpt,claude3} for describing inter-caption differences. Distinctively, our synthetic triplets provide multiple concise modification texts for each image pair, encompassing various attributes, in contrast to the conventional single, lengthy modification text. 
This enables a more nuanced representation of image modifications, potentially improving CIR model performance and versatility.

\mypara{Large Language Models.} Recent advancements in LLMs have expanded their applications beyond text generation to include image understanding~\cite{liu2023llava,liu2024llavanext} and text embedding generation~\cite{zhu2023large,mteb,wang2023improving}. Large Language Embedding Models (LLEMs) are specialized versions trained using contrastive learning, leveraging LLM knowledge for embedding generation while often disabling text generation capabilities. To enhance text embedding quality, \cite{lee2024nv,moreira2024nv,li2023towards} propose removing causal attention in LLMs, thereby improving text information encoding efficiency and training these modified LLMs on text retrieval datasets. This text retrieval capability has been further extended to text-image retrieval by aligning visual and LLM text embedding spaces~\cite{jang2024mate,mmgem,jiang2024e5}. These developments demonstrate the potential of LLMs for composed query understanding in CIR tasks. Our work builds upon this foundation, extending LLMs/LLEMs to generate composed query embeddings by incorporating reference images and modification text. This novel approach leverages the multimodal capabilities of LLMs to enhance the performance of CIR systems, leading to more accurate and context-aware image retrieval.

\begin{figure}[t]
    \centering
    \includegraphics[width=0.9\linewidth]{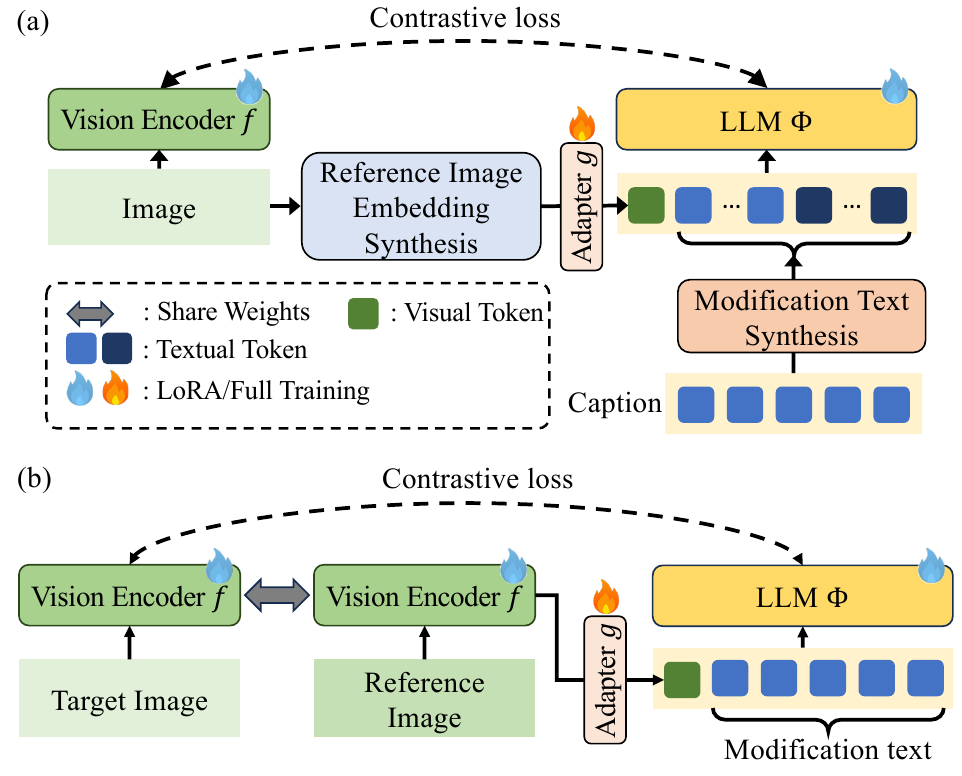}
    \vspace{-0.5em}
    \caption{An overview of our model and training strategies when using (a)~image-caption pairs and (b) CIR triplets. 
    }
    \label{fig:llm}
    \vspace{-1.5em}
\end{figure}

\vspace{-10pt}
\section{Method}
\label{sec:method}
This section outlines our methodology, starting with a brief overview of the model architecture, followed by a formal CIR problem definition. We then detail our triplet synthesis strategy, including reference image embedding and modification text synthesis, using image-caption pairs. Lastly, we describe our LLM-based query composition approach.

\vspace{-5pt}
\subsection{Model Architecture}
As illustrated in~\Fref{fig:llm}, CoLLM consists of several essential components: (1) a vision encoder $f(\cdot)$ for extracting image features; (2) modules for synthesizing reference image embeddings and modification text; (3) an image adapter $g(\cdot)$ that maps visual features into the language model's semantic space; (4) a LLM $\Phi(\cdot)$ that processes multimodal queries and (5) a projection layer $proj(\cdot)$ (omitted from the figure for simplicity) that maps the LLM output to a suitable representation for retrieval.
It is important to note that \Fref{fig:llm}~(a) and \Fref{fig:llm}~(b) illustrate architectures designed for two distinct input formats. The former is tailored for image-caption pairs, while the latter is optimized for CIR triplets.

\subsection{CoLLM with Image-Caption Pairs as Input}
\label{sec:method_pretrain}
Let $\mathcal{X} = \{(v_{i}, w_{i})\}^{N}_{i=1}$ denote a set of image-caption pairs, where $v_{i}$ and $w_{i}$ represent the image and caption of the $i^{th}$ sample, respectively. To effectively leverage the vast amount of image-caption pairs, we introduce a novel approach that synthesizes a CIR triplet on-the-fly for each image-caption pair $(v_{i}, w_{i})$. 
In this process, $v_{i}$ serves as the target image, while
two distinct modules generate the remaining components: (i) a reference image embedding synthesizer and (ii) a modification text synthesizer (\Fref{fig:llm}a).
Notably, we do not generate an actual image for reference image synthesis. Instead, we synthesize the reference image embedding, which is computationally more efficient and allows seamless integration into the CIR pipeline. This formulation enables CoLLM to exploit the rich information in image-caption pairs, transforming them into CIR triplets that can be used for training. The following subsections detail the specific mechanisms for synthesizing the reference image embedding and the modification text.

\begin{figure}[t]
    \centering
    \includegraphics[width=\linewidth]{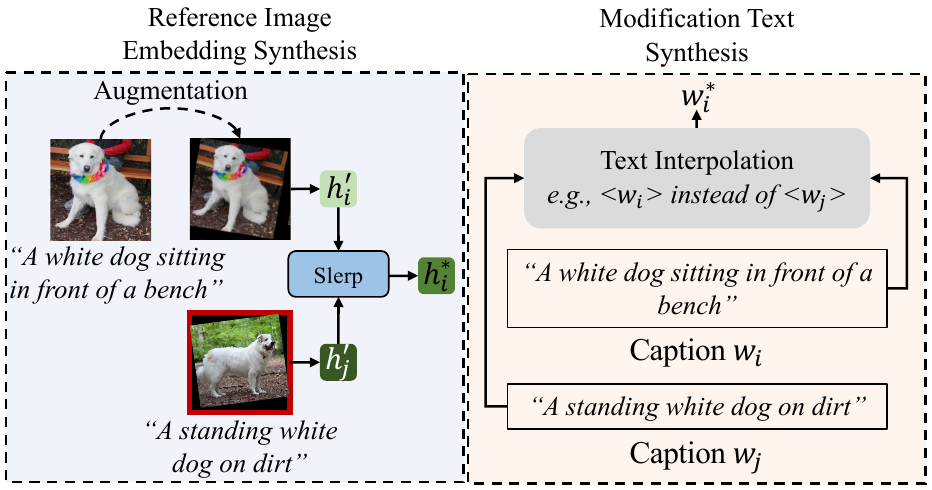}
    \vspace{-1em}
    \caption{An overview of reference image embedding synthesis and modification text synthesis. The red-framed image represents the nearest neighbor of the augmented image in the training batch.
    }
    \label{fig:synthesis}
    \vspace{-1.5em}
\end{figure}

\mypara{Reference Image Embedding Synthesis.}
Our reference image embedding synthesis process is illustrated in~\Fref{fig:synthesis} (left). We begin by applying an augmentation $t \sim \mathcal{T}$ to the input image $v_{i}$, where $\mathcal{T}$ represents a family of augmentations. The resulting augmented image is denoted as $v'_{i}$, with its corresponding embedding $\mathbf{h}'_{i}=f(v'_{i})$. For $\mathbf{h}'_{i}$, we identify its in-batch nearest neighbor ${v}'_j$ with $j$ defined as:
\begin{equation}
j = \argmax_{j \neq i ;\: j \in \{1,... \mathcal{B}\} } \text{sim}(\mathbf{h}'_{i}, \mathbf{h}'_j),
\end{equation}
where $\mathcal{B}$ is the batch size, $\mathbf{h}'_j=f(v'_j)$, and $\text{sim}(\mathbf{u}, \mathbf{v})={\mathbf{u}^{T} \mathbf{v}} / ({\|\mathbf{u}\| \|\mathbf{v}\|})$ denotes cosine similarity.

We then employ Spherical Linear Interpolation (Slerp)~\cite{slerp} to synthesize the reference image embedding by interpolating between $\mathbf{h}'_{i}$ and $\mathbf{h}'_j$. The synthesized reference image embedding is obtained as:
\begin{equation}
    \begin{aligned}
        \theta &= \arccos (\mathbf{h}'_{i} \cdot \mathbf{h}'_j) \\
        \mathbf{h}_{i}^{\ast} &= \frac{\sin(\alpha\theta)}{\sin{\theta}} \mathbf{h}'_{i} + \frac{\sin\left((1-\alpha)\theta\right)}{\sin\theta} \mathbf{h}'_j,
    \end{aligned}
\end{equation}
where $\alpha \in [0, 1]$ is a hyper-parameter controlling the interpolation strength. The intuition behind this approach is that an image with embedding $\mathbf{h}_{i}^{\ast}$ shares certain visual similarities with the target image $v_{i}$. Specifically, a larger $\alpha$ results in $\mathbf{h}_{i}^{\ast}$ more closely resembling $\mathbf{h}'_i$, allowing for fine-grained control over the synthesized embedding.

\mypara{Modification Text Synthesis.}
For the augmented image $v'_{i}$, its nearest neighbor is $(v'_{j}, w_{j})$. To generate the modification text $w_{i}^{\ast}$, we use text interpolation to combine $w_{i}$ and $w_{j}$ using pre-defined templates, as illustrated in \Fref{fig:synthesis} (right). The complete set of templates is provided in the supplementary material. During training, we randomly select a template for each sample to ensure diversity in the synthesized modification text. It is worth noting that we do not simply use $w_{i}$ as the modification text for two primary reasons: (i) $w_{i}$ alone fails to capture the visual differences between the reference image and the target image. (ii) Using only $w_{i}$ could lead to model cheating, where it learns to rely solely on $w_{i}$ for retrieval while ignoring the reference image.
Our synthesis approach, in contrast, mimics real-world modification text, which often describes both similarities and differences between the reference and target images. Furthermore, this strategy forces the model to learn composed query embeddings by considering both the reference image and the modification text simultaneously.

By combining the synthesized reference embedding $\mathbf{h}_{i}^{\ast}$ and the interpolated modification text $w_{i}^{\ast}$, our method generates diverse CIR triplets from image-caption pairs. This approach enables effective training of the CoLLM model on large-scale image-caption datasets, effectively mitigating the dependency on scarce labeled triplet data.

\mypara{Query Composition with LLM.}
To construct the composed query, we utilize a pre-trained LLM that processes the synthesized reference image embedding $\mathbf{h}_{i}^{\ast}$, image caption $w_i$,
and the synthesized modification text $w_{i}^{\ast}$. We define three distinct composed embeddings:
\begin{align}
    \mathbf{c}_i^{v} &= p(\Phi(g(\mathbf{h}_{i}^{\ast}))) \label{eq:ref} \\
    \mathbf{c}_i^{w} &= p(\Phi(w_{i})) \label{eq:mod} \\
    \mathbf{c}_i &= p(\Phi([g(\mathbf{h}_{i}^{\ast}); w_{i}^{\ast}])) \label{eq:ref_mod} 
\end{align}
where $[;]$ denotes an instruction template that combine two modalities (see supplementary material).

\mypara{Training Objective.}
We employ a contrastive loss $\mathcal{L}_{cl}$~\cite{radenovic2023filtering}, consistent with previous works~\cite{webcovr,thawakar2024composed}, to compute the loss between query embeddings $\{\mathbf{c}_i^{v}, \mathbf{c}_i^{w}, \mathbf{c}_i \}$ and the target image embedding $\mathbf{z}_i=f(v_i)$. The final loss for each sample during pre-training is defined as:
\begin{equation}
    \mathcal{L} = \frac{1}{3} \left( \mathcal{L}_{cl}(\mathbf{c}_i^{v},\mathbf{z}_i) + \mathcal{L}_{cl}(\mathbf{c}_i^{w},\mathbf{z}_i) + \mathcal{L}_{cl}(\mathbf{c}_i,\mathbf{z}_i) \right)\label{eq:loss}
\end{equation}
This formulation encourages the model to learn discriminative representations for different query types while maintaining consistency with the target image embedding. By combining losses from image-to-image ($\mathbf{c}_i^{v}$), text-to-image ($\mathbf{c}_i^{w}$), and composed ($\mathbf{c}_i$) queries, we ensure that the model learns to effectively process and align various input modalities for compositional image retrieval.

\begin{table*}[t!]
    \centering
    \vspace{-1.5em}
    \caption{
    Comparison of synthetic CIR training datasets.} %
    \vspace{-6pt}
    \footnotesize
    \begin{tblr}{width=\textwidth,colsep=3pt,colspec={l|c|cccc|cccc},row{9}={mypink}}
    \toprule
    \SetCell[r=2]{l}  Dataset & \SetCell[r=2]{c} Public & \SetCell[c=4]{c} Reference-Target Image Pair & & & & \SetCell[c=4]{c} Modification Text Generation \\
    \hline
    & & Type & Source & \# Pairs & \# Entities & LLM & LLM input & \# Text & \# Text/Pair \\
    \hline
    LaSCo~\cite{caselasco} & \checkmark & image & VQAv2~\cite{goyal2017vqav2} & 360K & 82K & GPT-3~\cite{gpt3} & question-answer & 360K & 1 \\
    VDG~\cite{vdg} & \checkmark &  image & NLVR2~\cite{suhr2019corpus}, COCO~\cite{lin2014microsoft} & 467K & 183K & Vision-LLaMA2 & image pair & 523K & 1.12 \\
    WebCoVR~\cite{webcovr} & \checkmark & video &  WebVid2M~\cite{webvid2m} & 1.6M & 131K & Custom LLM & caption & 1.6M & 1 \\
    CC-CoIR~\cite{ventura24covr2} & \checkmark & image & CC3M~\cite{sharma2018conceptual} & 3.3M & 357K & Custom LLM & caption & 3.3M & 1 \\
    SynthTriplets~\cite{compodiff} & & synth. image & SD~\cite{rombach2022high}, IP2P~\cite{brooks2023instructpix2pix} & 18.8M & 37.6M & OPT-6.7B~\cite{zhang2022opt} & - & 18.8M & 1 \\
    MagicLens~\cite{magiclens} &  & image & Web crawled & 36.5M & - & PaLM~\cite{chowdhery2023palm} & image metadata & 36.5M & 1 \\
    MTCIR (Ours) & \checkmark & image & LLaVA-558k~\cite{liu2023llava} & 3.4M & 423K & Sonnet 3~\cite{claude3} & synth. caption & 17.7M & 5.18 \\
    \bottomrule
    \end{tblr}
    \label{tab:dataset}
    \vspace{-1em}
\end{table*}

\subsection{CoLLM with CIR Triplet as Input}
For CIR triplet inputs, our model design is shown in \Fref{fig:llm}b. Let $\mathcal{X} = \{(v_{i}^{r}, v_{i}^{t}, w_{i})\}^{N}_{i=1}$ be a set of triplets, where $v_{i}^{r}$, $v_{i}^{t}$, and $w_{i}$ represent the reference image, target image, and modification text of the $i^{th}$ sample, respectively.
The reference embedding is obtained as $\mathbf{h}_{i} = f(v_{i}^{r})$. The composed embedding is then computed as $\mathbf{c}_i = p(\Phi([g(\mathbf{h}_{i}); w_{i}]))$.
The training objective is $\mathcal{L} = \mathcal{L}_{cl}(\mathbf{c}_i,\mathbf{z}_i)$, where $\mathbf{z}_i=f(v_{i}^{t})$ denotes the target embedding.

\begin{figure}[t]
    \centering
    \vspace{-5pt}
    \includegraphics[width=0.85\linewidth]{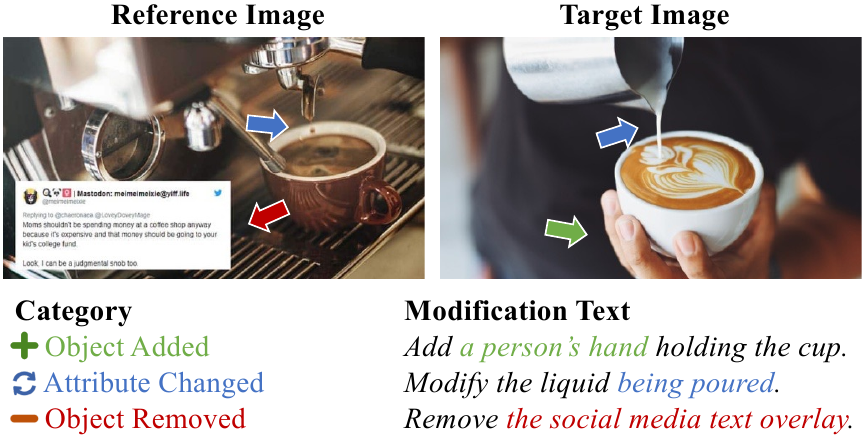}
    \vspace{-0.5em}
    \caption{An example from the MTCIR dataset. Each sample contains multiple short texts describing different modifications.}
    \label{fig:mtcir_examples}
    \vspace{-1.5em}
\end{figure}

\vspace{-5pt}
\section{Dataset Construction}
\label{sec:dataset}
This section outlines the data construction process for two primary objectives: (1) creating a novel CIR training dataset called Multi-Text CIR (MTCIR) and (2) refining widely used CIR benchmarks, specifically CIRR~\cite{cirr} and Fashion-IQ~\cite{fashioniq}. We detail the methodologies for each task, highlighting the improvements and innovations.

\subsection{Multi-Text CIR (MTCIR) Dataset}
\label{sec:train_dataset}
Existing CIR triplet datasets~\cite{caselasco,vdg,webcovr} lack diversity and contain unnatural modification text. We introduce Multi-Text CIR (MTCIR) to address these limitations. 

To enhance image diversity, we source images from the LLaVA-558k dataset \cite{liu2023llava} filtered from image caption datasets~\cite{sharma2018conceptual,schuhmann2021laion,sbu} based on noun-phrase frequency for broad concept coverage. We pair images using CLIP \cite{clip} visual similarity metrics, following CIRR's grouping approach, yielding 3.4M pairs from approximately 423K images. Leveraging the detailed captions generated by LLaVA-Next-34B \cite{liu2024llavanext} for each image in the LLaVA-558k dataset, we employ Claude 3 Sonnet \cite{claude3} (\texttt{20240229-v1:0}) to generate modification text for each image pair, using these captions as input.
To control detail levels in Claude outputs, we define six modification text categories in the prompt, inspired by prior research \cite{searle,cirr}. This approach maintains diverse aspects and prevents repetition. We also incorporate examples from the CIRR dataset to ensure human-like modification text. Detailed prompting strategies are provided in the supplementary material.

Our method generates multiple brief, focused texts for each image pair, each highlighting a unique aspect. This approach captures all changes between two images without producing overly long sentences, better aligning with human-written modification text. These short texts can be combined for training, enhancing dataset diversity. We obtained 17.7M text samples for the 3.4M image pairs, averaging 5.18 short sentences per pair. \Fref{fig:mtcir_examples} illustrates an example, showcasing modification texts corresponding to specific categories and covering diverse attributes.

\begin{figure}[t]
    \centering
    \vspace{-5pt}
    \includegraphics[width=\linewidth]{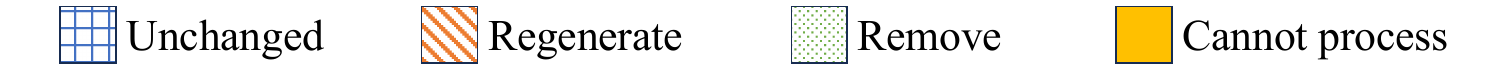}
    \begin{subfigure}[b]{0.24\linewidth}
        \includegraphics[width=\linewidth]{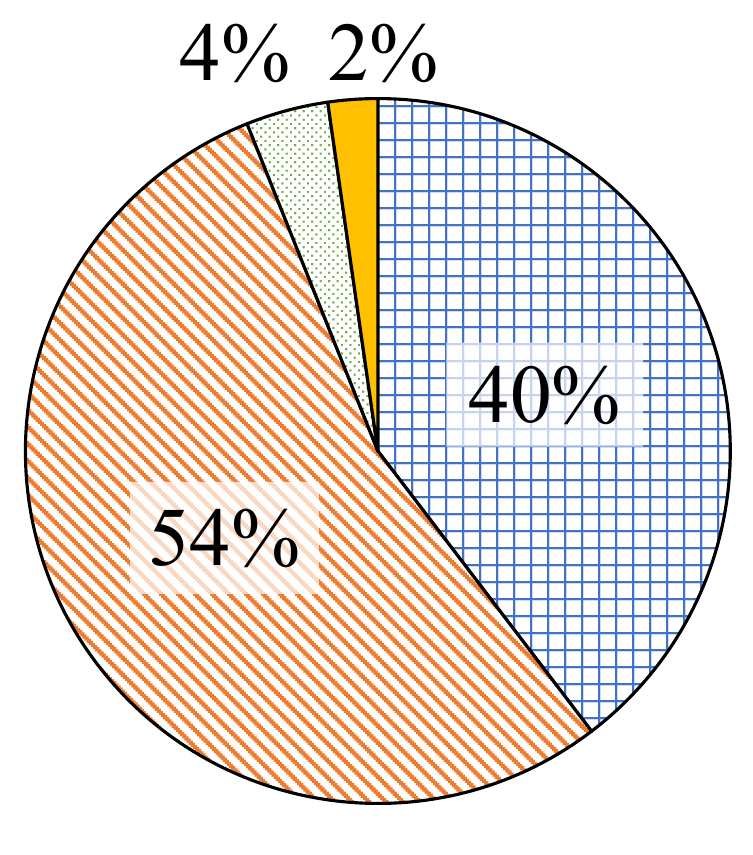}
        \caption{CIRR}
    \end{subfigure}
    \begin{subfigure}[b]{0.24\linewidth}
        \includegraphics[width=\linewidth]{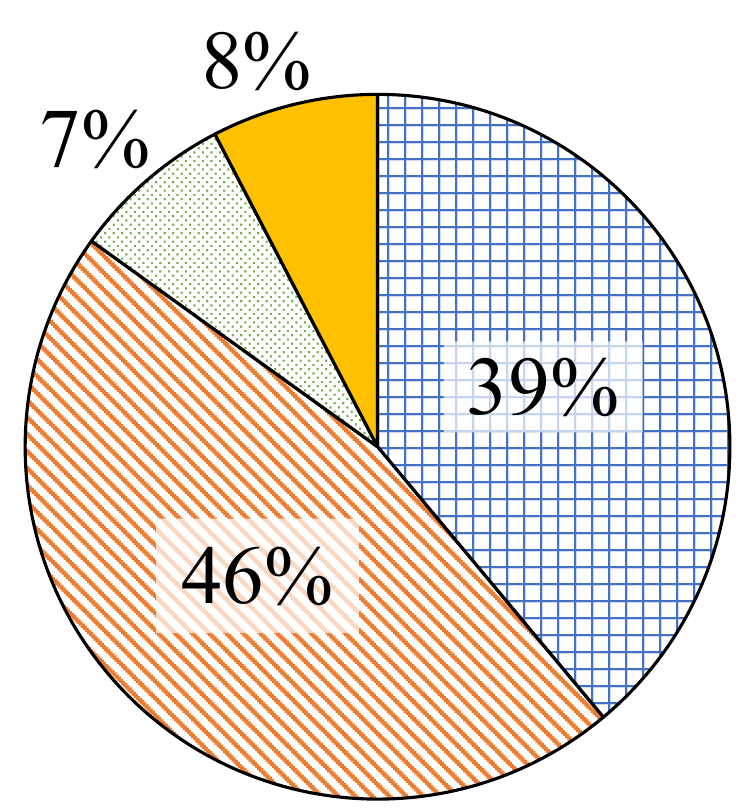}
        \caption{FIQ-Dress}
    \end{subfigure}
    \begin{subfigure}[b]{0.24\linewidth}
        \includegraphics[width=\linewidth]{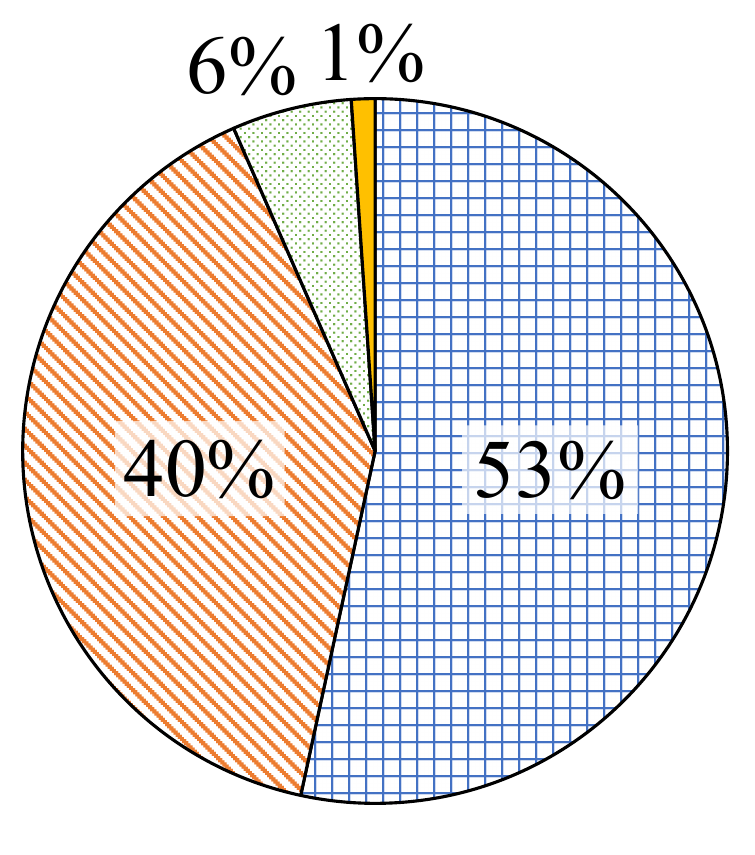}
        \caption{FIQ-Shirt}
    \end{subfigure}
    \begin{subfigure}[b]{0.24\linewidth}
        \includegraphics[width=\linewidth]{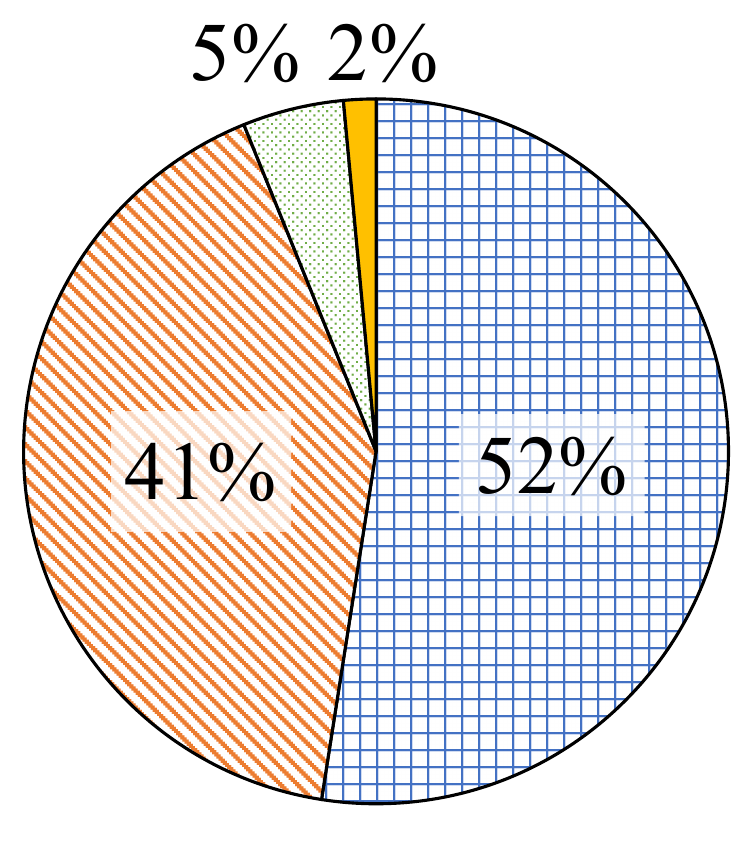}
        \caption{FIQ-Toptee}
    \end{subfigure}
    \caption{Statistical analysis of the refinement process for CIRR and FashionIQ (FIQ) datasets.}
    \label{fig:benchmark_results}
    \vspace{-5pt}
\end{figure}

\begin{figure}[t!]
    \centering
    \includegraphics[width=0.85\linewidth]{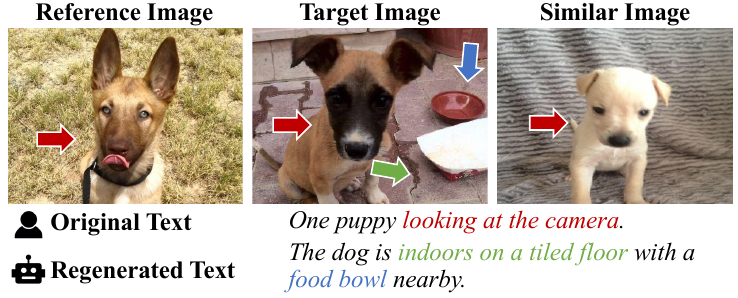}
    \vspace{-0.5em}
    \caption{
    An example from the refined CIRR dataset. The original modification text yields multiple plausible results (shown as similar images to the target on the right). In contrast, the newly generated text accurately captures the distinctive features of the target image, differentiating it from other candidates.}
    \label{fig:benchmark_samples}
    \vspace{-1.5em}
\end{figure}

As shown in \Tref{tab:dataset}, MTCIR offers several advantages over existing synthetic CIR datasets: (i) Unlike \cite{compodiff}, which uses synthetic images, MTCIR uses real images covering diverse concepts and domains. (ii) MTCIR provides multiple modification texts for each image pair, offering more natural descriptions and training flexibility. (iii) MTCIR 
is the largest public dataset to advance CIR research.

To prevent the leakage of biometric information, we further process MTCIR at both the image and text levels. For images, we apply face blurring to all instances. For text, we remove any content containing keywords related to human attributes (e.g., skin, hair, gender, age, race).

\subsection{Refined CIRR and Fashion-IQ Datasets}
\label{sec:new_benchmarks}
Current CIR benchmarks often suffer from label ambiguity, where modification text lacks clarity, potentially matching multiple target images. To address this issue, we propose a refinement method using Claude 3 Sonnet \cite{claude3} (\texttt{20240229-v1:0}) to enhance the validation sets of two well-known benchmarks: CIRR and Fashion-IQ. Each sample in these validation sets comprises a triplet containing a reference image, target image, and modification text.

\begin{table}[t!]
    \centering
    \vspace{-1.5em}
    \caption{Performance comparison of various models and visual encoders on CIR benchmarks, evaluated without training on triplet datasets. \textbf{Bold} and \underline{underlined} values denote the best and second-best scores within each vision encoder group. Models incorporating LLMs in their architectures are marked with $^\star$. Results reproduced by our team are indicated by $^\ddagger$.}
    \vspace{-6pt}
    \footnotesize
    \begin{tblr}{width=\linewidth,colspec={@{}X[4,l]|X[1,c]X[1,c]X[1,c]|X[1,c]X[1,c]X[1,c]|X[1,c]X[1,c]@{}},row{3,9,18}={myorange},row{8,17,20}={mypink},stretch = 0,colsep=2pt}
    \toprule
         \SetCell[r=2]{c} Method
         & \SetCell[c=3]{c} CIRCO (mAP$\uparrow$) & &
            & \SetCell[c=3]{c} CIRR (Rec.$\uparrow$) & &
                & \SetCell[c=2]{c} FIQ (Rec.$\uparrow$) \\
    \hline
    & @5 & @10 & @50 &@1 &@10 & @50 & @10 & @50 \\
    \hline
    \SetCell[c=9]{c} OpenAI CLIP-B/32  \\
    PALAVRA~\cite{palavra} & 4.6 & 5.3 & 6.8 & 16.6 & 58.5 & 84.0 & 19.8 & 37.3  \\ 
    SEARLE~\cite{searle} & 9.4 & 9.9 & 11.8 & 24.0 & 66.8 & \snd{89.8} & 22.9 & 42.5  \\
    Slerp-TAT~\cite{slerptat} & 9.3 & 10.3 & 12.3 & \snd{28.2} & \snd{68.8} & 88.5 & 23.0 & 44.0  \\ 
    CIReVL$^\star$~\cite{cirevl} & \fst{14.9} & \fst{15.4} & \fst{17.8} & 23.9 & 66.0 & 87.0 & \fst{28.3} & \fst{49.4}  \\
    CoLLM$^\star$ & \snd{12.9} & \snd{13.2} & \snd{15.0} & \fst{28.6} & \fst{71.8} & \fst{92.7} & \snd{24.8} & \snd{45.2} \\
    \hline
    \SetCell[c=9]{c} OpenAI CLIP-L/14  \\
    Pic2World~\cite{pic2word} & 8.7 & 9.5 & 11.3 & 23.9 & 65.3 & 87.8 & 24.7 & 43.7  \\ 
    SEARLE~\cite{searle} & 11.7 & 12.7 & 15.1 & 24.2 & 66.3 & 88.8 & 25.6 & 46.2  \\ 
    LinCIR$^\ddagger$~\cite{lincir} & 13.0 & 13.9 & 16.2 & 24.6 & 66.9 & 88.8 & 26.4 & 46.6  \\ 
    ContextI2W~\cite{contexti2w} & 13.0 & 13.8 & 16.0 & 25.7 & 68.6 & 89.6 & 27.8 & 48.9  \\ 
    MCL$^\star$~\cite{mcl} & 17.8 & 18.4 & 21.8 & 25.9 & 69.4 & \snd{91.0} & - & -  \\ 
    Slerp-TAT~\cite{slerptat} & 18.5 & \snd{19.4} & \snd{21.4} & \fst{30.9} & \snd{70.9} & 89.2 & 28.3 & 47.6  \\ 
    CIReVL$^\star$~\cite{cirevl} & \snd{18.6} & 19.0 & 21.8 & 24.6 & 64.9 & 86.3 & \snd{28.6} & \snd{48.6}  \\ 
    CoLLM$^\star$ & \fst{20.3} & \fst{20.8} & \fst{23.4} & \snd{29.7} & \fst{72.8} & \fst{91.5} & \fst{30.1} & \fst{49.5}\\
    \hline
    \SetCell[c=9]{c} BLIP-L/16  \\
    Slerp-TAT~\cite{slerptat} & \snd{17.8} & \snd{18.4} & \snd{21.1} & \snd{34.0} & \snd{72.7} & \snd{88.9} & \snd{32.8} & \snd{53.3} \\
    CoLLM$^\star$ & \fst{19.7} & \fst{20.4} & \fst{23.1} & \fst{35.0} & \fst{78.6} & \fst{94.2} & \fst{34.6} & \fst{56.0} \\ %
    \bottomrule
    \end{tblr}
    \label{tab:pretrain}
    \vspace{-15pt}
\end{table}

Our refinement process, applicable to any benchmark, consists of validation, re-generation, and re-validation steps. For each triplet, we first identify the top-3 visually similar images to the target image based on CLIP visual scores, serving as hard negative samples for ambiguity assessment. We employ Claude 3 Sonnet to evaluate each triplet's ambiguity by attempting to identify the target image from these hard negatives. Triplets correctly identified by Claude 3 Sonnet are considered ``good" samples and remain unchanged. For ``bad" samples, we use Claude 3 Sonnet to re-generate the modification text. The model produces three new modification texts with increasing detail in a hierarchical structure, with each finer text building upon the coarser one. 
We then repeat the validation process, evaluating the new texts from coarsest to finest and selecting the best one. Any re-generated modification text that passes the validation process replaces the original text and concludes the process. If all generated texts fail the validation, the triplet is removed from the benchmark. Detailed process information is provided in the supplementary material. Besides, following MTCIR, biometric information is also removed from the refined benchmarks. 

As illustrated in \Fref{fig:benchmark_results}, up to 8\% of samples are omitted due to Claude's processing limitations related to harmful content. Additionally, 4-7\% of ambiguous samples that could not be effectively rewritten are removed. More than 40\% of the original samples are retained, and 40-55\% of triplets are successfully revised.
The refined CIRR and Fashion-IQ datasets exhibit less ambiguity (\Fref{fig:benchmark_samples}), offering more robust evaluation benchmarks for the CIR community.

\begin{table}[t!]
    \centering
    \vspace{-1.5em}
    \caption{Performance of models trained on synthetic triplet datasets. \textbf{Bold} indicates the best method overall, while \underline{underline} highlights the best method within the same vision encoder. Our CoLLM outperforms comparable methods in most metrics.}
    \vspace{-6pt}
    \footnotesize
    \begin{tblr}{width=\linewidth,colsep=4pt,colspec={@{}X[5,l]|X[6,l]|X[1,c]X[1,c]X[1,c]|X[1,c]X[1,c]},row{10,14}={mypink},row{3,5,7,11}={myorange},stretch = 0}
    \toprule
         \SetCell[r=2]{l} Method
         & \SetCell[r=2]{l}{Dataset}
            & \SetCell[c=3]{c} CIRR $\uparrow$ & &
                & \SetCell[c=2]{c} FIQ $\uparrow$ \\
    \hline
    & & @1 & @10 & @50 & @10 & @50 \\
    \hline
    \SetCell[c=7]{c} CoCa-L/18 $288\x 288$ \\
    MagicLens~\cite{magiclens} & MagicLens~\cite{magiclens}  & 33.3 & 77.9 & 94.4 & 38.1 & 58.3\\ 
    \hline
    \SetCell[c=7]{c} EVA-CLIP ViT-G/14 $364\x364$ \\
     CoVR2~\cite{ventura24covr2} & WV-CC-CoVIR~\cite{ventura24covr2} & 43.7 & 84.0 & \fst{96.1} & 38.2 & 58.4 \\
    \hline
    \SetCell[c=7]{c} OpenAI CLIP-L/14 $224\x 224$ \\
    CompoDiff~\cite{compodiff} & SynTrip18M~\cite{compodiff} & 18.2 & 70.8 & 90.3 & \snd{36.0} & 48.6\\ 
    MagicLens~\cite{magiclens} &  MagicLens~\cite{magiclens}  & 30.1 & 74.4 & 92.6 & 30.7 & 52.5\\ 
    CoLLM & MTCIR (ours) & \snd{34.7} & \snd{77.0} & \snd{93.1} & 32.9 & \snd{54.2} \\
    \hline
    \SetCell[c=7]{c} BLIP-L/16 $384\x 384$; fine-tuned on COCO \\
    CASE~\cite{caselasco} & LaSCo~\cite{caselasco} & 35.4 & 78.5 & 94.6 & - & -\\ 
    Omkar et al.~\cite{thawakar2024composed} & WebCoVR~\cite{webcovr} & 40.1 & 78.9 & 94.7 & 30.3 & 46.5\\ 
    CoLLM & MTCIR (ours) & \fst{\snd{45.8}} & \fst{\snd{84.7}} & \snd{95.8} & \fst{\snd{39.1}} & \fst{\snd{60.7}} \\ %
    \bottomrule
    \end{tblr}
    \label{tab:ft}
    \vspace{-15pt}
\end{table}

\section{Experiments}
\label{sec:exp}
Our proposed CoLLM framework adopts a two-stage training paradigm: pre-training and fine-tuning phases. In pre-training, CoLLM is exposed to a diverse corpus of image-caption pairs, enabling it to understand composed queries comprehensively. Subsequently, in the fine-tuning stage, we leverage our newly curated MTCIR datasets to verify their effective performance on complex multimodal tasks.

\subsection{Pre-Training on Image-Caption Pairs}
\label{sec:exp_pretrain}
\mypara{Training and Evaluation Datasets.} The pre-training dataset consists of 5 million image-caption pairs, compiled from CC3M \cite{sharma2018conceptual}, LAION \cite{schuhmann2021laion}, and LLaVA-558K \cite{liu2023llava}. We evaluate our model on three CIR benchmarks: CIRCO \cite{searle}, CIRR \cite{cirr} test set, and Fashion-IQ \cite{fashioniq} validation set. All datasets are filtered to exclude biometric information.

\mypara{Evaluation Metrics.} We employ recall on the complete image index set as the evaluation metric for CIRR and Fashion-IQ. For CIRCO, we use mean Average Precision (mAP). We exclude the recall on subset for CIRR due to reliability concerns (see supplementary material for details).

\mypara{Implementation Details.} See supplementary material.

\mypara{Quantitative Results.}~\Tref{tab:pretrain} compares our pre-trained model with other methods not trained on triplet datasets. We evaluate three variants of our architecture, each employing different vision encoders.
CoLLM with the CLIP-B encoder achieves the second-best performance on the CIRCO and Fashion-IQ benchmarks while significantly improving on the CIRR benchmark. Although CIReVL~\cite{cirevl} outperforms other methods, its use of the closed-source GPT-4~\cite{achiam2023gpt} introduces additional costs and increased inference latency, making it an unfair comparison.

For larger vision encoders, our model consistently outperforms both non-LLM (e.g., Slerp-TAT~\cite{slerptat}) and LLM  (CIReVL and MCL~\cite{mcl}) methods across all benchmarks. 
Notably, our CLIP-L variant achieves state-of-the-art results on the CIRCO benchmark, while BLIP-L significantly improves CIRR and Fashion-IQ scores. 
The consistent gains across benchmarks and architectures validate the robustness and generalizability of our triplet synthesis approach. By effectively bridging the gap between abundant image-caption data and scarce triplet annotations, our method paves the way for more scalable and efficient training of CIR models.
This innovative strategy represents a significant step forward in addressing one of the critical challenges in the field, Paving the way for larger-scale pre-training and more complex multimodal interactions.

\begin{table}[t!]
    \centering
    \vspace{-1.5em}
    \caption{Performance of BLIP-L and CoLLM (pre-trained) trained on synthetic triplet datasets. \textbf{Bold} values indicate the best score within each group. Models trained on MTCIR demonstrate superior performance compared to those trained on other datasets.}
    \vspace{-6pt}
    \footnotesize
    \begin{tblr}{width=\linewidth,colspec={@{}X[5,l]|X[6,l]|X[1,c]X[1,c]X[1,c]|X[1,c]X[1,c]},row{5,7}={mypink},stretch = 0}
    \toprule
         \SetCell[r=2]{l} Method
         & \SetCell[r=2]{l}{Dataset}
            & \SetCell[c=3]{c} CIRR $\uparrow$ & &
                & \SetCell[c=2]{c} FIQ $\uparrow$ \\
    \hline
    & & @1 & @10 & @50 & @10 & @50 \\
    \hline
     \SetCell[r=3]{l} BLIP-L~\cite{blip} & LaSCo~\cite{caselasco} & 36.6 & 78.4 & 94.6 & 24.8 & 44.0\\ 
     & WebCoVR~\cite{webcovr} & 39.3 & 78.9 & 94.7 & 26.7 & 43.3\\ 
     & MTCIR (ours) & \fst{42.4} & \fst{83.1} & \fst{95.9} & \fst{37.9} & \fst{59.2} \\
     \hline
     \SetCell[r=2]{l} CoLLM & LaSCo~\cite{caselasco} & 43.2 & 84.1 & 95.7 & 38.5 & 60.1 \\
     & MTCIR (ours) & \fst{45.8} & \fst{84.7} & \fst{95.9} & \fst{39.1} & \fst{60.7} \\ 
    \bottomrule
    \end{tblr}
    \label{tab:ft_data}
    \vspace{-15pt}
\end{table}

\subsection{Fine-tuning on MTCIR}
\label{sec:exp_ft}
\mypara{Training Dataset and Benchmarks.} Despite the impressive performance achieved using only image-caption datasets, we hypothesize that CoLLM's performance can be further enhanced through fine-tuning on our newly developed MTCIR dataset. For a fair comparison, we also train CoLLM and baseline models on other public datasets, including WebCoVR~\cite{webcovr} and LaSCo~\cite{caselasco}. Our evaluation focuses on the CIRR~\cite{cirr} test and Fashion-IQ~\cite{fashioniq} validation sets. We exclude CIRCO~\cite{searle} due to potential domain overlap between COCO~\cite{lin2014microsoft} images used in CIRCO and those in LaSCo~\cite{caselasco}, as well as in the pre-training datasets.

\mypara{Implementation Details.} See supplementary material.

\mypara{Quantitative Results.}~\Tref{tab:ft} presents the evaluation of models trained on publicly available synthetic CIR datasets, including our MTCIR. 
Fine-tuning CoLLM on MTCIR yields significant gains, improving CIRR@1 by 5\% over CoLLM-OpenAI-CLIP-L/14 in~\Tref{tab:pretrain}.
This improvement reflects MTCIR's enhanced generalizability and our superior data construction. Our method achieves exceptional results on all benchmarks using the same vision encoder compared to previous approaches and models trained on other synthetic datasets. Notably, our fine-tuned CoLLM outperforms MagicLens CoCa-L~\cite{magiclens} and CoVR2~\cite{ventura24covr2}, despite these models using larger, more advanced vision encoders and being trained on larger datasets. These results underscore the importance of data quality over quantity.

\begin{table}[t!]
    \centering
    \vspace{-1.5em}
    \caption{
    Performance of models on refined CIRR and Fashion-IQ validation sets. \textbf{Bold} indicates the highest score, while \underline{underlined} values represent the best metric within the same vision encoder group. CoLLM+MTCIR continues to outperform other configurations on these new benchmarks.}
    \vspace{-6pt}
    \footnotesize
    \begin{tblr}{width=\linewidth,colspec={@{}X[5,l]|X[6,l]|X[1,c]X[1,c]X[1,c]|X[1,c]X[1,c]},row{7,11}={mypink},row{3,5,8}={myorange},stretch = 0}
    \toprule
         \SetCell[r=2]{l} Method
         & \SetCell[r=2]{l}{Dataset}
            & \SetCell[c=3]{c} Ref. CIRR $\uparrow$ & &
                & \SetCell[c=2]{c} Ref. FIQ $\uparrow$ \\
    \hline
    & & @1 & @10 & @50 & @10 & @50 \\
    \hline
    \SetCell[c=7]{c} EVA-CLIP ViT-G/14 $364\x364$ \\
     CoVR2~\cite{ventura24covr2} & WV-CC-VIR~\cite{ventura24covr2} & 56.1 & 91.1 & 97.9 & 54.2 & 72.9  \\
    \hline
    \SetCell[c=7]{c} OpenAI CLIP-L/16 $224\x 224$ \\
    MagicLens~\cite{magiclens} &  MagicLens~\cite{magiclens}  & 42.3 & 86.3 & \snd{97.0} & 45.5 & 68.1\\ 
    CoLLM & MTCIR (ours) & \snd{46.5} & \snd{87.4} & 96.1 & \snd{48.3} & \snd{68.6} \\
    \hline
    \SetCell[c=7]{c} BLIP-L/16 $384\x 384$; fine-tuned on COCO \\
    BLIP-L~\cite{blip} & MTCIR (ours) & 53.8 & 89.9 & 97.7 & 54.8 & 74.3 \\
    CoLLM & LaSCo~\cite{caselasco} & 57.3 & 92.0 & 98.1 & 56.9 & 75.9 \\ 
    CoLLM & MTCIR (ours) & \fst{\snd{60.4}} & \fst{\snd{92.6}} & \fst{\snd{98.2}} & \fst{\snd{57.2}} & \fst{\snd{76.4}} \\ %
    \bottomrule
    \end{tblr}
    \label{tab:reeval}
    \vspace{-15pt}
\end{table}

\mypara{MTCIR vs. Other CIR Datasets.} Despite CoLLM's state-of-the-art performance, two questions remain: (i) Can MTCIR benefit other models? (ii) Can CoLLM achieve good performance when fine-tuned on other CIR datasets? To address these questions, we train a baseline BLIP-L~\cite{blip} model on various CIR datasets, including MTCIR. Both LaSCo and MTCIR experiments are trained for one epoch. As shown in Table~\ref{tab:ft_data}, we observe that: (i) BLIP-L shows significant improvement when using MTCIR as the training data, and (ii) CoLLM exhibits notable performance degradation when fine-tuned on the LaSCo dataset.
These results demonstrate MTCIR's versatility as a plug-and-play dataset, capable of enhancing performance across various models. This underscores MTCIR's potential to make significant contributions to the CIR community.

\vspace{-5pt}
\subsection{Re-evaluation on Refined Benchmarks}
\label{sec:exp_reeval}

Ambiguity in the CIRR and Fashion-IQ benchmarks potentially skews the metrics presented in previous tables, particularly for high-performing models. While CIRCO mitigates this issue, it still suffers from domain overlap with training images. To assess the impact of ambiguous samples, we re-evaluate the models on our newly refined CIRR and Fashion-IQ benchmarks. \Tref{tab:reeval} presents the performance of these models on the revised benchmarks.

Our analysis shows that model rankings remain largely consistent before and after the benchmark refinement. As the modification texts in the refined benchmarks provide additional nuance, we anticipated that models with more robust composed query understanding capabilities would distinguish themselves. Indeed, CoLLM consistently outperforms other models across various settings on the refined benchmarks.
Notably, while BLIP-L (MTCIR) trailed CoLLM (MTCIR) by 3.4\% (CIRR@1)(~\Tref{tab:ft_data}), this performance gap widened to 6.9\% (CIRR@1) on the refined CIRR benchmark. This suggests CoLLM better captures nuanced modification text and distinguishes visual differences between source and target images.

\vspace{-5pt}
\subsection{Ablation Studies}
\label{sec:exp_abl}

\begin{table}[t!]
    \centering
    \vspace{-1.5em}
    \caption{
    Ablation studies of proposed components.
    }
    \vspace{-6pt}
    \begin{subtable}[t]{0.85\linewidth}
        \centering
        \caption{Reference image and modification text interpolation. 
        }
        \label{tab:abl_syn}
        \footnotesize
        \begin{tblr}{width=\linewidth,colspec={X[1.3,c]X[1.3,c]|X[2,c]X[2,c]},stretch = 0,colsep=1pt}
            \toprule
                Image & Text & CIRCO (mAP $\uparrow$) & CIRR (Rec. $\uparrow$) \\
            \hline
            & & 14.7 & 170.1 \\
            & \checkmark & 14.5 & 176.4 \\
            \checkmark & & 39.6 & 183.3 \\ 
            \checkmark & \checkmark & \textbf{52.8} & \textbf{194.5} \\
            \bottomrule
        \end{tblr}
        \vspace{6pt}
    \end{subtable}
    \begin{subtable}[t]{0.85\linewidth}
        \centering
        \caption{Unimodal queries in \Eref{eq:ref} and \Eref{eq:mod}. 
        }
        \label{tab:abl_query}
        \footnotesize
        \begin{tblr}{width=\linewidth,colspec={X[1.3,c]X[1.3,c]|X[2,c]X[2,c]},stretch = 0,colsep=1pt}
            \toprule
                Image-only & Text-only  & CIRCO (mAP $\uparrow$) & CIRR (Rec. $\uparrow$) \\
            \hline
            \checkmark &  & 14.5 & 99.7 \\
              & \checkmark & 47.1 & \textbf{196.3} \\
            \checkmark & \checkmark & \textbf{52.8} & 194.5 \\
            \bottomrule
        \end{tblr}
        \vspace{6pt}
    \end{subtable}
    \begin{subtable}[t]{0.85\linewidth}
        \centering
        \caption{Reference image embedding interpolation.
        }
        \label{tab:abl_slerp}
        \footnotesize
        \begin{tblr}{width=\linewidth,colspec={@{}X[1.75,l]|X[1.1,c]X[1.1,c]},stretch = 0,colsep=1pt}
            \toprule
                 & CIRCO (mAP $\uparrow$) & CIRR (Rec. $\uparrow$) \\
            \hline
            Random In-Batch Sample & 46.7 & 182.2 \\ 
            Nearest In-Batch Neighbor & \textbf{52.8} & \textbf{194.5} \\
            \bottomrule
        \end{tblr}
    \end{subtable}
    \label{tab:abl}
    \vspace{-15pt}
\end{table}

We conduct comprehensive ablation studies to evaluate the impact of our proposed components during the pre-training stage. Our experiments use the BLIP-L encoder, trained exclusively on the LLaVA-558k~\cite{liu2023llava} dataset for one epoch. \Tref{tab:abl} presents the performance across various settings. We report the sum of mAP at $k=\{5,10,25,50\}$ for CIRCO and Recall@$k=\{1,5,10,50\}$ for CIRR. \textbf{Image and Text Interpolation are Crucial:} As shown in \Tref{tab:abl_syn}, employing Slerp for reference image embedding significantly enhances the model's learning capabilities. Applying interpolation on modification text further improves performance. \textbf{Unimodal Queries are Beneficial:} We investigate the necessity of unimodal queries in \Eref{eq:ref} and \Eref{eq:mod} and observe that omitting unimodal queries during training, especially text-only queries, substantially degrades performance (\Tref{tab:abl_query}). \textbf{Nearest Neighbor is Essential:} For reference image embedding interpolation, one might hypothesize that a random in-batch sample could suffice for interpolation. However, our study demonstrates that using the nearest image embedding neighbor yields significantly better performance (\Tref{tab:abl_slerp}). \textbf{LLEMs Outperforms LLMs:} We investigate Large Language Embedding Models (LLEMs), which are fine-tuned for text retrieval compared to their base LLMs. As shown in \Tref{tab:abl_llm}, both E5-Mistral-7B-Inst \cite{wang2023improving} and SFR-Embedding-2 outperform their LLM counterpart, Mistral-7B \cite{jiang2023mistral}. The superior performance of LLEMs demonstrates that models specifically tailored for embedding and retrieval tasks can offer substantial advantages in CIR applications compared to general-purpose language models.

\begin{table}[t!]
    \centering
    \vspace{-1.5em}
    \caption{Performance of different LLMs with CLIP-L/14 as the visual encoder. \textbf{Bold} and \underline{underline} highlight the best and second best score.
    $^\star$ indicates the original LLM for text generation; others are Large Language Embedding Models (LLEMs) fine-tuned for text retrieval.
    }
    \vspace{-6pt}
    \footnotesize
    \begin{tblr}{width=\linewidth,colspec={@{}X[5,l]|X[1,c]X[1,c]X[1,c]|X[1,c]X[1,c]X[1,c]|X[1,c]X[1,c]},row{6}={mypink},stretch = 0,colsep=1.5pt}
    \toprule
         \SetCell[r=2]{c} LLM
         & \SetCell[c=3]{c} CIRCO (mAP$\uparrow$) & &
            & \SetCell[c=3]{c} CIRR (Rec.$\uparrow$) & &
                & \SetCell[c=2]{c} FIQ (Rec.$\uparrow$) \\
    \hline
    & @5 & @10 & @50 &@1 &@10 & @50 & @10 & @50 \\
    \hline
    Stella-Qwen2-1.5B~\cite{stellaqwen2} & 14.8 & 15.3 & 17.4 & 27.5 & 69.1 & 88.0 & 29.3 & 49.0 \\
    Mistral-7B$^\star$~\cite{jiang2023mistral} & \snd{19.8} & 20.2 & 22.7 & \snd{29.6} & 72.5 & \snd{91.3} & 29.5 & \snd{49.3} \\
    E5-Mistral-7B-Inst~\cite{wang2023improving} & 19.6 & \snd{20.3} & \snd{22.8} & 29.5 & \snd{72.7} & 91.1 & \snd{29.9} & \fst{49.5} \\
    SFR-Embedding-2~\cite{SFREmb2} & \fst{20.3} & \fst{20.8} & \fst{23.4} & \fst{29.7} & \fst{72.8} & \fst{91.5} & \fst{30.1} & \fst{49.5}\\
    \bottomrule
    \end{tblr}
    \label{tab:abl_llm}
    \vspace{-15pt}
\end{table}

\vspace{-10pt}
\section{Limitations and Conclusion}

While these advancements significantly improve CIR performance, several areas warrant further investigation. Our work, limited to LLM/LLEMs, could benefit from exploring pre-trained MLLMs for enhanced understanding of CIR tasks. Additionally, our triplet synthesis method generates a single visual token, constraining the model's ability to process detailed image information. Future work should leverage the underutilized text category information in our synthetic datasets to improve model generalization. Lastly, our refined benchmarks, while more reliable, still contain noise from original image pairs, suggesting a need for further refinement of evaluation metrics.

In conclusion, we present novel approaches to Composed Image Retrieval (CIR) that obviate the need for annotated datasets. Our contributions include: (1) an innovative triplet synthesis method utilizing image-caption pairs, (2) a new architecture leveraging LLM's embedding generation capabilities, (3) MTCIR, a diverse, human-aligned synthetic dataset, and (4) refined versions of CIRR and Fashion-IQ benchmarks, enhancing the reliability of evaluation metrics in the field. Our method consistently outperforms existing LLM and non-LLM baselines across popular benchmarks. Notably, MTCIR achieves superior results with only 10-20\% of the size of larger datasets, demonstrating a 1-15\% increase in Recall. These advancements collectively push the boundaries of CIR, offering more efficient and effective solutions for real-world applications.

{
    \small 
    \bibliographystyle{ieeenat_fullname}
    \bibliography{main}
}

\clearpage
\maketitlesupplementary

\section{Additional Method Details}

\subsection{Modification Text Synthesis Templates}

As described in~\Sref{sec:method_pretrain} and illustrated in \Fref{fig:synthesis} (right), the synthesis of modification text plays a vital role in the initial pre-training stage. During this stage, we generate modification text $w_i^\ast$ by randomly choosing one of the templates provided below:

\begin{enumerate}
    \item ``show $w_i$ instead of $w_j$"
    \item ``$w_i$ instead of $w_j$"
    \item  ``show $w_i$ rather than $w_j$"
    \item  ``$w_i$ rather than $w_j$"
    \item  ``rather than $w_j$, show $w_i$"
    \item  ``rather than $w_j$, $w_i$"
    \item  ``instead of $w_j$, $w_i$"
    \item ``$w_j$, changed to $w_i$"
    \item ``not $w_j$, but $w_i$"
    \item ``show $w_i$, not $w_j$"
    \item ``$w_j$ is missing, $w_i$"
    \item  ``$w_i$, and $w_j$ is missing"
    \item  ``remove $w_j$, add $w_i$"
    \item  ``add $w_i$, remove $w_j$"
    \item  ``$w_j$ become $w_i$"
\end{enumerate}
The templates are designed based on our analysis of the real modification texts from the CIRCO and CIRR datasets, aiming to integrate information from both the reference and target images. While the fully synthesized modification texts may not be grammatically or semantically correct, the language encoder is pre-trained to handle such noise robustly.

\subsection{LLM Instruction Template}
\label{sec:instruction_template}
As stated in~\Eref{eq:ref}-(\ref{eq:ref_mod}), the input to the LLM must adhere to a specific template. We adopt the LLEM (LLM specialized for text retrieval) instruction format to structure our input instruction as:

\begin{verbatim}
Instruct: Find the image that matches 
          the query.
Query:
Image: [IMAGE]
Text: [TEXT]
\end{verbatim}

\noindent where \texttt{[IMAGE]} corresponds to $g(\mathbf{h_i^\ast})$ or $g(\mathbf{h_i})$, and \texttt{[TEXT]} corresponds to $w_i^\ast$ or $w_i$ when training with image-caption pairs or triplets, respectively. If either \texttt{[IMAGE]} or \texttt{[TEXT]} is missing, the line \texttt{Image: [IMAGE]} or \texttt{Text: [TEXT]} is removed from the query accordingly.

\subsection{Additional Ablation Studies}

\begin{figure}[t!]
    \centering
    \includegraphics[width=0.6\linewidth]{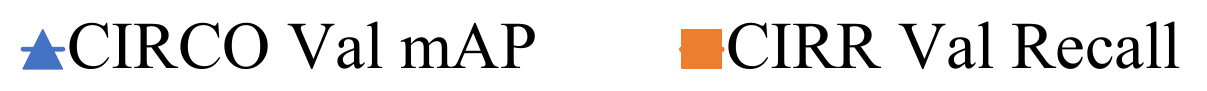}
    \begin{subfigure}[b]{0.49\linewidth}
        \includegraphics[width=\linewidth]{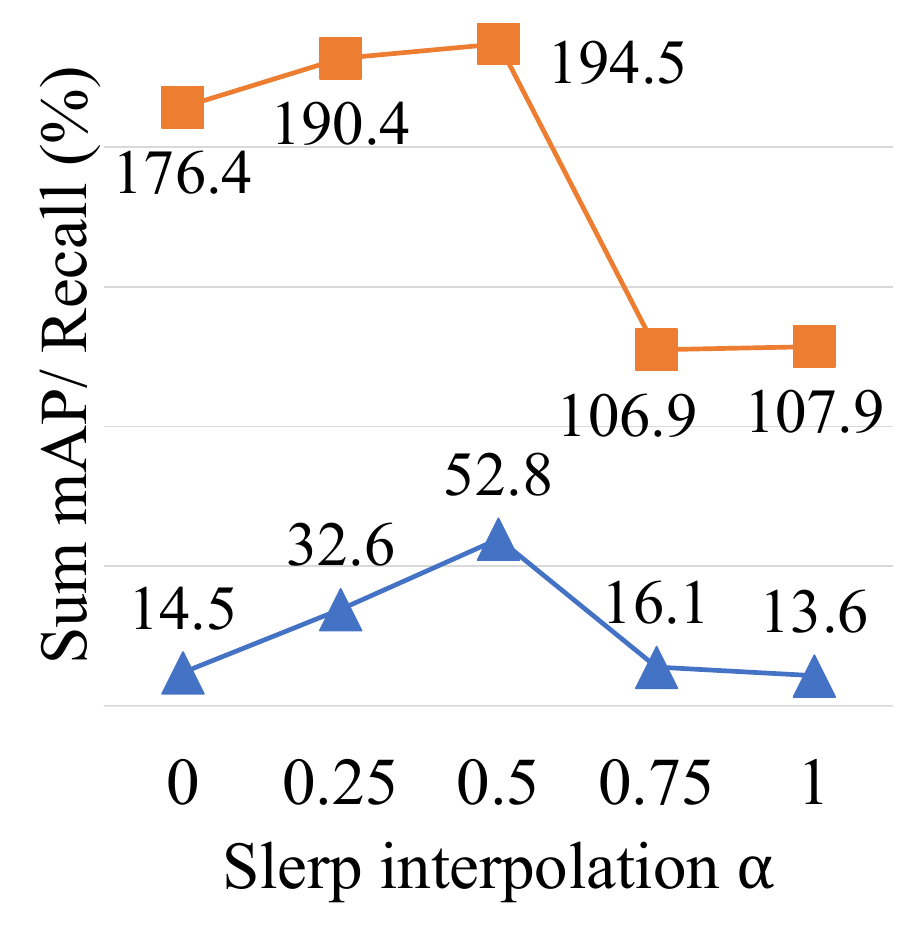}
        \caption{Different Slerp $\alpha$ values}
        \label{fig:pretrain_abl_alpha}
    \end{subfigure}
    \begin{subfigure}[b]{0.49\linewidth}
        \includegraphics[width=\linewidth]{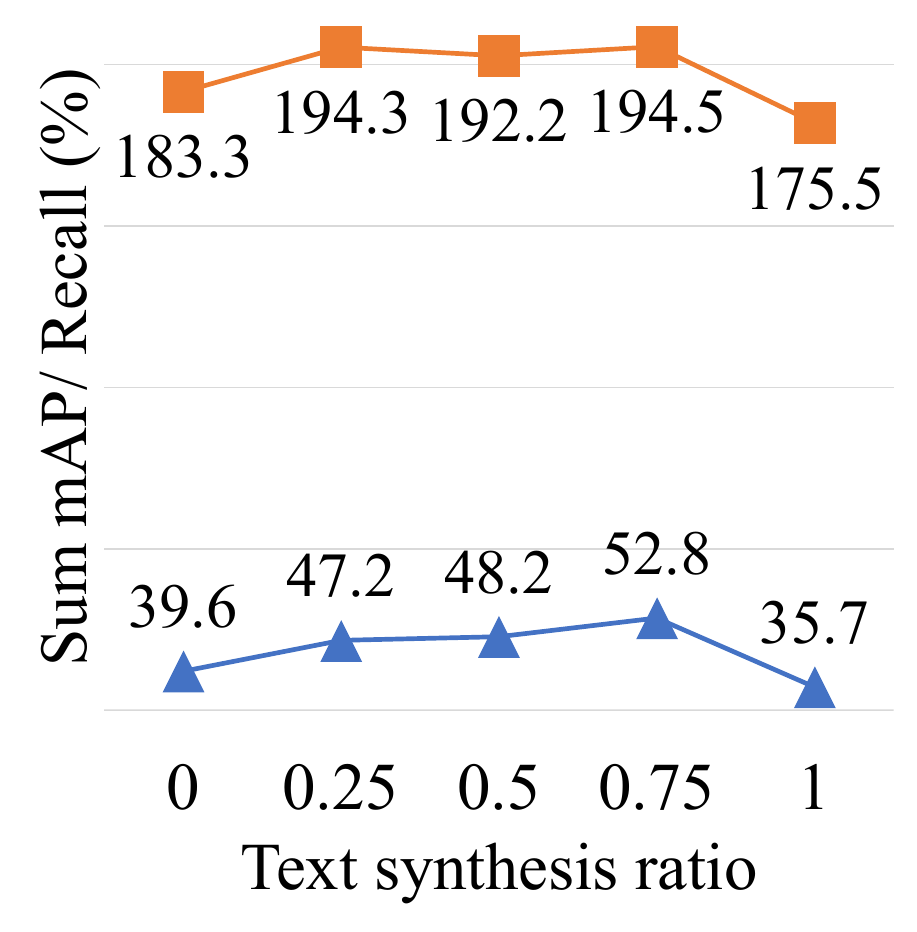}
        \caption{Different Text synthesis ratios}
        \label{fig:pretrain_abl_text}
    \end{subfigure}
    \caption{Performance of our model under varying Slerp $\alpha$ values and text synthesis ratios. Text synthesis = 75\% in (a) and Slerp $\alpha=0.5$ in (b). The optimal configuration is achieved with $\alpha=0.5$, where text synthesis is applied to 75\% of the samples. }
    \label{fig:abl_syn_strength}
\end{figure}

\begin{table}[t!]
    \centering
    \caption{Performance of our CoLLM BLIP-L/16 ($384\x 384$) finetuned on COCO after training on MTCIR and test with different instructions on CIRR Val and Fashion-IQ.}
    \footnotesize
    \begin{tblr}{width=\linewidth,colspec={@{}X[l]|X[c]X[c]X[c]|X[c]X[c]@{}},stretch = 0}
    \toprule
    \SetCell[r=2]{c}
    & \SetCell[c=3]{c} CIRR Val & &
    & \SetCell[c=2]{c} FIQ\\
    \hline
    &@1 &@10 & @50 & @10 & @50 \\
    \hline
    Mean & 47.0 & 85.7 & 96.0 & 38.7 & 60.6\\
    Std & 0.10 & 0.04 & 0.04 & 1.02 & 0.48 \\
    \bottomrule
    \end{tblr}
    \label{tab:abl_instruction}
\end{table}

We investigate the impact of synthesis strength hyperparameters for the reference image embedding $\mathbf{h_i^\ast}$ and modification text $w_i^\ast$ in \Fref{fig:abl_syn_strength}. The same training setup as described in~\Sref{sec:exp_abl} is used. As explained in~\Sref{sec:method_pretrain}, the Slerp $\alpha$ value represents the interpolation distance of the reference image embedding relative to the original $\mathbf{h}'_i$. A larger $\alpha$ value indicates a greater difference between $\mathbf{h_i^\ast}$ and $\mathbf{h_j^\ast}$. For modification text synthesis, it is applied partially to the training samples. When synthesis does not occur, $w_i^\ast=w_i$, the caption of the target image. From the figures, the model achieves optimal performance with Slerp $\alpha=0.5$ and text synthesis applied to 75\% of the training samples. Performance drops significantly with higher $\alpha$ values.

To assess the robustness of our model across different instructions, we generate nine additional instruction variants using Claude Sonnet, as described in~\Sref{sec:instruction_template}:
\begin{enumerate}
    \item ``Identify the image corresponding to the given query."
    \item ``Locate the image that aligns with the provided query."
    \item ``Search for the image that fits the query."
    \item ``Retrieve the image that matches the query."
    \item ``Determine the image that corresponds to the query."
    \item ``Select the image that best matches the query."
    \item ``Find the image associated with the query."
    \item ``Choose the image that matches the given query."
    \item ``Match the query to its corresponding image."
\end{enumerate}

As shown in~\Tref{tab:abl_instruction}, when tested with ten different instructions, our model demonstrates robustness, exhibiting negligible performance variation across instruction variants.

\section{Dataset Construction Details}

\subsection{MTCIR}
\mypara{Image Pairing.} The process follows CIRR~\cite{cirr} with some modifications. Specifically, we use CLIP-L-14/336~\cite{clip} to extract image features instead of ResNet-152~\cite{resnet} pre-trained on ImageNet~\cite{deng2009imagenet}. This updated network provides more robust features compared to the previous one. Groups of six similar images are formed, where each image is added to the group with a similarity score between 0.5 and 0.95 relative to the first image, using an interval of 0.03. Groups with fewer than six members are discarded. Pairs are then constructed between consecutive images and between the first image and all other images within each group.

\mypara{Modification Text Categories.} We define six categories as outlined in~\Tref{tab:category}, drawing inspiration from previous works, CIRR~\cite{cirr} and CIRCO~\cite{searle}. The largest category, Attribute Changed, comprises approximately half of the dataset's text. Object Added and Object Removed have similar proportions, each accounting for around 20\% of the dataset. The remaining three categories collectively represent less than 10\% of the dataset.

\begin{table*}[t!]
    \centering
    \vspace{-1em}
    \caption{Modification text categories define six types of changes that can occur between two images. These categories capture the variety of transformations described in the dataset.}
    \footnotesize
    \begin{tblr}{width=\linewidth,colspec={@{}X[1.2,l]X[1,l]X[1,c]X[3,l]@{}},stretch = 0}
    \toprule
    Category ID & Name & No. Samples & Definition \\
    \hline
    \texttt{attribute\_change} & Attribute Changed &  8,139,415 (45.95\%) & The same object is present in both images, but the attributes of the object have changed, not including the quantity or number. \\
    \texttt{added\_object} & Object Added & 3,856,642 (21.77\%) & An object or objects is present in the second image that is not present in the first image. \\
    \texttt{removed\_object} & Object Removed & 3,695,121 (20.86\%) & An object or objects is present in the first image that is not present in the second image. \\
    \texttt{relationship\_change} & Relationship Changed & 1,122,834 (6.34\%) & If the objects in the images are the same, but the relationship between the objects has changed. \\
    \texttt{viewpoint\_change} & Viewpoint Changed & 650,735 (3.67\%) & The viewpoint from which the image is taken has changed between the two images. \\
    \texttt{number\_change} & Number Changed & 249,098 (1.41\%) & The same object is present in both images, but the number of the object has changed. \\
    \bottomrule
    \end{tblr}
    \label{tab:category}
    \vspace{-0.5em}
\end{table*}

\mypara{Prompt.} The input to Claude Sonnet 3 is detailed in~\Tref{tab:prompt}. It begins with a system prompt that provides an overview of the task to the model. Next, the detailed image captions (\texttt{[CAPTION]}) generated by LLaVA-Next-34B~\cite{liu2024llavanext} are included, followed by the definitions of categories outlined in~\Tref{tab:category}.

For each category, real captions and modification texts from CIRR~\cite{searle} (with some corrections) are provided as examples to enable in-context learning. Both forward examples (\texttt{[FORWARD]}: describing changes from image 1 to image 2) and backward examples (\texttt{[BACKWARD]}: describing changes from image 2 to image 1) are included to ensure the model accurately understands the task.

Additionally, during the initial iterations, a set of bad examples (\texttt{[BAD EXAMPLES]}), which fail to describe the changes correctly, is collected and incorporated into the prompt to refine the model's understanding. Finally, the JSON output requirement is specified at the end for straightforward parsing.

This prompt structure allows for the generation of multiple modification texts in both forward and backward directions for a single image pair, reducing costs while ensuring all detailed differences are captured.

We include 20 MTCIR samples in \url{mtcir_samples.html} of the supplementary material, showcasing various modification texts and categories for each. Our pipeline effectively captures the differences between pairs without producing lengthy sentences.

\mypara{Diversity and Quality.} We evaluate the diversity of the MTCIR dataset by analyzing variability in both image content and textual descriptions. For image diversity, we utilize RAM~\cite{zhang2024recognize} to tag images in our dataset and those in previously published benchmarks. For textual diversity, we employ spaCy~\cite{honnibal2020spacy} to process modification texts. As presented in~\Tref{tab:diversity}, our dataset achieves the highest count of unique visual tags and textual tokens, indicating superior diversity.

To assess dataset quality, we employ state-of-the-art LLMs to evaluate the generated modification texts. Specifically, we use GPT-4o~\cite{hurst2024gpt} and DeepSeek-V3~\cite{liu2024deepseek}, two leading-performing models, to validate the accuracy of modifications. Each model is provided with captions of the reference and target images alongside our generated modification text and tasked to identify any incorrect transformations described by the modification text. The evaluation is conducted on 1000 randomly selected samples from the MTCIR dataset. Our dataset achieves good sample ratios of 83.4\% and 85.2\%, as rated by GPT-4o and DeepSeek-V3, respectively.

\begin{table}[h!]
    \centering
    \caption{MTCIR is more diverse than previous datasets in both visual and textual information.}
    \footnotesize
    \begin{tblr}{width=\linewidth,colspec={@{}lrrrr@{}},stretch=0,colsep=2.6pt}
    \toprule
    & CIRR~\cite{cirr} & LaSCo~\cite{caselasco} & CC-CoIR~\cite{ventura24covr2} & MTCIR \\
    \hline
    \# Unique Visual Tags & 2,787 & 3,421 & 4,072 & \textbf{4,198} \\
    \# Unique Text Tokens & 5,838 & 16,270 & 18,031 & \textbf{164,914} \\
    \bottomrule
    \end{tblr}
    \label{tab:diversity}
\end{table}

\subsection{Refined Benchmarks}
\begin{figure}[t]
    \centering
    \vspace{-1em}
    \includegraphics[width=\linewidth]{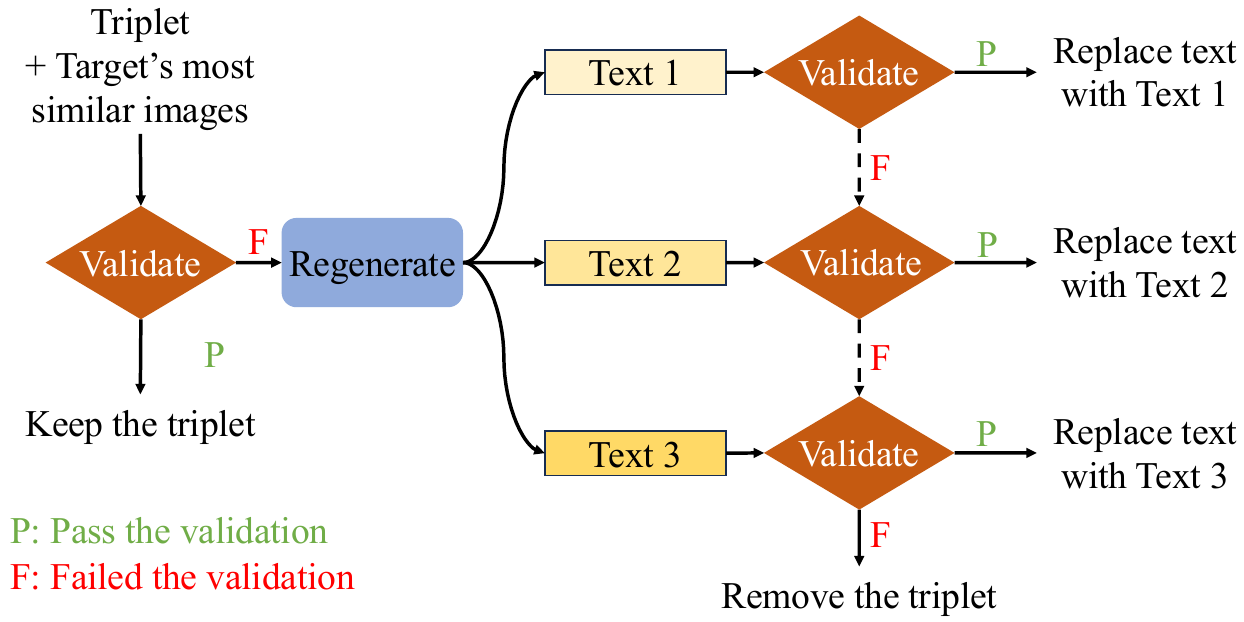}
    \caption{\textbf{Pipeline for regenerating text in CIR benchmarks.} Starting with a triplet and dataset-similar images, we assess text ambiguity by evaluating the model's ability to select the correct target image. If the model fails, three new texts with varying levels of specificity are generated and re-tested. The process concludes either when ambiguity is resolved by any of the texts or when the triplet is removed if ambiguity remains unresolved.}
    \label{fig:benchmark_pipeline}
    \vspace{-1.5em}
\end{figure}

The refinement pipeline, as detailed in~\Sref{sec:new_benchmarks}, is illustrated in~\Fref{fig:benchmark_pipeline}. It consists of three steps to ensure that only ``good" samples remain in the benchmarks.

In the first and final steps, sample validation is conducted using the prompt outlined in~\Tref{tab:benchmark_prompt_validate}. The reference image is included as \texttt{[REFERENCE IMAGE]}, while the target image and all hard negative samples (the top-3 similar images to the target image) are concatenated horizontally in random order as \texttt{[CANDIDATE IMAGES]}. Given the modification text as \texttt{[MODIFICATION TEXT]}, the Claude Sonnet model is tasked with selecting the correct target image. Each sample is evaluated three times with different orders of \texttt{[CANDIDATE IMAGES]}. Samples that pass in at least two evaluations are considered ``good." Occasionally, the model refuses to answer, providing responses beginning with ``I apologize...", a behavior triggered by its harmful content detection mechanism. Such samples are excluded from the benchmarks.

In the second step, modification texts for ambiguous samples are regenerated. Claude Sonnet 3 is used to create new modification texts, guided by the prompt described in~\Tref{tab:benchmark_prompt_regen}. The original triplet is retained as input with \texttt{[REFERENCE IMAGE]}, \texttt{[TARGET IMAGE]}, and \texttt{[MODIFICATION TEXT]}. Additionally, some randomly selected ``good" samples from the first step are included as \texttt{[GOOD SAMPLES]} to align the model's output with human expectations. The prompt instructs the model to generate three modification texts, ranging from coarse to fine, to minimize inference costs.

\begin{figure*}
    \centering
    \includegraphics[width=\linewidth]{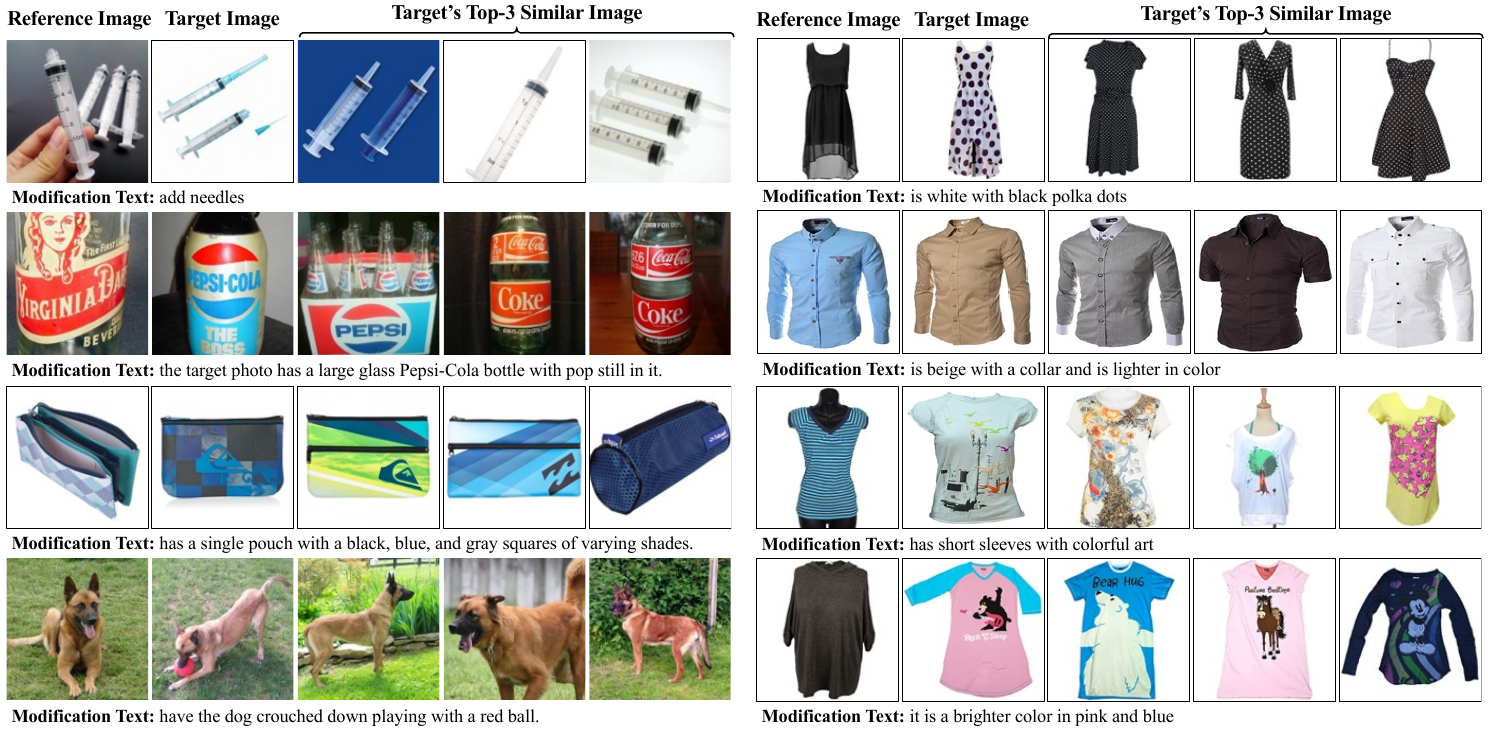}
    \caption{``Good" samples kept in the Refined CIRR (left) and Fashion-IQ (right). The original modification text correctly highlights the different between target and most similar images.}
    \label{fig:benchmark_good}
\end{figure*}

\begin{figure*}
    \centering
    \includegraphics[width=\linewidth]{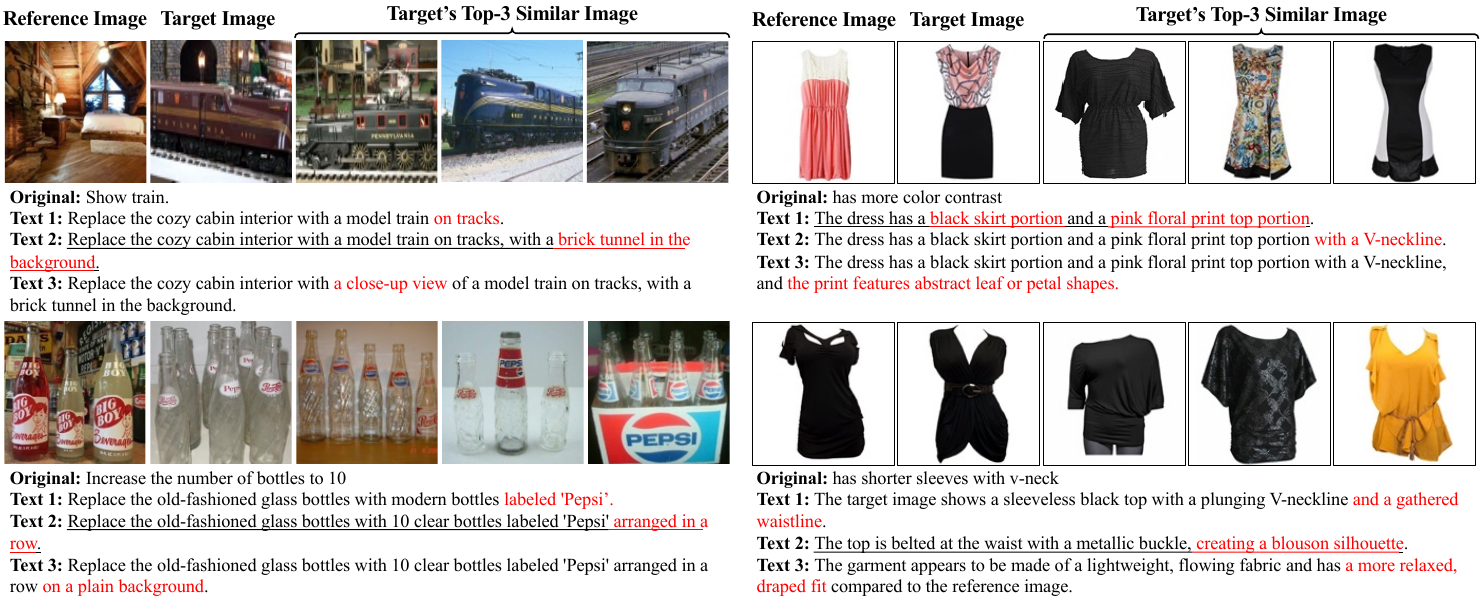}
    \includegraphics[width=\linewidth]{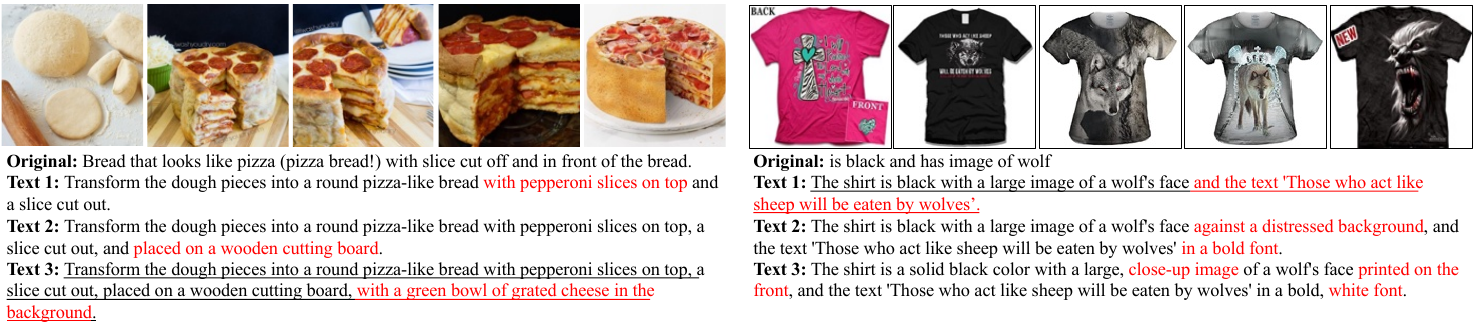}
    \caption{``Bad" samples with re-generated text in the Refined CIRR (left) and Fashion-IQ (right). The \underline{underlined} is the selected modification text to replace the original. \textcolor{red}{Red} highlights the adding detail from the original to finest Text.}
    \label{fig:benchmark_bad}
\end{figure*}

We present some ``good" samples classified by our pipeline from both the CIRR and Fashion-IQ validation sets in~\Fref{fig:benchmark_good}. The modification texts in these samples are sufficiently detailed to distinguish the target image from hard-negative samples, which are visually similar to the target. Examples of Text 1-3 are shown in~\Fref{fig:benchmark_bad} along with new chosen text. In these examples, our pipeline prioritizes using coarse modification text to replace ambiguous ones. At each level, an additional detail is introduced to further differentiate the correct target image from the other hard-negative samples.

\section{Additional Experimental Details}
\subsection{Pre-training on Image-Caption Pairs.}

\mypara{CIRR Recall on Subset Metric.} During our evaluation on the CIRR validation set, we observed some contradictions between Recall on the whole index set and Recall on the subset. These inconsistencies raise concerns about the reliability of the recall on the subset metric. We evaluate BLIP-L baselines with the vision encoder BLIP-L/16 fine-tuned on COCO, using different settings and the synthetic CIR dataset, as shown in~\Tref{tab:sub_metric}. While our proposed MTCIR achieves the best results, some interesting observations emerge regarding Recall Subset.

Firstly, the initial model already outperforms the fine-tuned models trained on previous synthetic datasets. Notably, this model uses only modification text in the query and achieves the second-best performance, with a small gap to the best-performing model. Additionally, the model fine-tuned on WebCoVR shows slight degradation in performance when both image and text are used in the query. These results suggest that the reference image does not play a significant role in the Recall Subset, indicating that this metric is unreliable for evaluating CIR methods.

\begin{table}[t!]
    \centering
    \caption{Unreliability of Recall Subset metric on CIRR validation. The BLIP-L/16 384 ft. COCO model is trained on various CIR datasets and evaluated under different query settings. Notably, even without fine-tuning, the initial model achieves the second-best performance, surpassing all previous datasets while not using the reference image in the query. \textbf{Bold} and \underline{underline} are used to highlight the best and second-best scores, respectively.}
    \footnotesize
    \begin{tblr}{width=\linewidth,colsep=1pt,colspec={@{}X[3,l]|X[1,c]X[1,c]|X[1,c]X[1,c]X[1,c]|X[1,c]X[1,c]X[1,c]@{}},stretch = 0}
         \toprule
         \SetCell[r=2]{l}Fine-tuned Dataset 
            & \SetCell[c=2]{c}Query & 
            & \SetCell[c=3]{c}Recall Index $\uparrow$ & &
            & \SetCell[c=3]{c}Recall SubSet $\uparrow$ \\
         \hline
         & Image & Text & @1 & @10 & @50 & @1 & @2 &@3 \\
         \hline
         \SetCell[r=2]{l} None & & \checkmark & 38.5 & 75.1 & 89.3 & \snd{75.5} & \snd{88.4} & \snd{94.2} \\
         & \checkmark & \checkmark & 20.6 & 54.7 & 76.2 & 67.0 & 84.5 & 91.7 \\
         \hline
         \SetCell[r=2]{l}LaSCo~\cite{caselasco} & & \checkmark & 23.4 & 59.2 & 80.0 & 65.1 & 82.9 & 91.0 \\
         & \checkmark & \checkmark & 40.4 & 80.9 & 94.9 & 68.2 & 84.0 & 91.5 \\
         \hline
         \SetCell[r=2]{l}WebCoVR~\cite{webcovr} & & \checkmark & 34.3 & 73.0 & 88.7 & 73.3 & 87.5 & 93.7 \\
         & \checkmark & \checkmark & \snd{40.6} & \snd{81.5} & \snd{94.5} & 72.7 & 87.4 & 93.5 \\
         \hline
         \SetCell[r=2]{l}MTCIR & & \checkmark & 22.2 & 58.1 & 79.4 & 67.3 & 84.9 & 92.9 \\
         & \checkmark & \checkmark & \fst{43.9} & \fst{84.1} & \fst{95.4} & \fst{75.6} & \fst{89.3} & \fst{95.4} \\
         \bottomrule
    \end{tblr}
    \label{tab:sub_metric}
\end{table}

\mypara{Image-Caption Datasets.} We provide additional details about the pre-training dataset mentioned in the main paper. Our dataset, comprising 5 million image pairs, follows the Slerp-TAT~\cite{slerptat} protocol: nearly 3 million samples are sourced from CC3M~\cite{sharma2018conceptual}, 2 million random samples are selected from LAION-115M~\cite{schuhmann2021laion}, and 558K samples are obtained from LLaVA-558K~\cite{liu2023llava}. All captions are synthetically generated using the BLIP~\cite{blip} model.

\mypara{Implementation Details.} We utilize SFR-Embedding-2~\cite{SFREmb2} as our LLM backbone. For the vision encoder, we employ pre-trained OpenAI CLIP-B/32 and CLIP-L/14~\cite{clip}. We adopt BLIP-L/16 pre-trained weights from the official repository~\cite{blip}. To ensure a fair comparison, all pre-training experiments use $224 \times 224$ pixel images.

LoRA~\cite{lora} tuning is applied with a rank of 64 for large models (BLIP-L and CLIP-L) and 32 for the CLIP-B model, using a dropout rate of 0.1. The BLIP-L vision encoder is frozen during training, while other model variants are tuned on both the LLM and vision encoder parts.

Pre-training is conducted over one epoch using a constant learning rate of $1e^{-4}$ and a batch size of 1024. All experiments are performed on 8 NVIDIA A100 40GB GPUs. The training script is based on the LLaVA~\cite{liu2023llava} codebase, while the evaluation script is adopted from WebCoVR~\cite{webcovr}.

\mypara{Different BLIP Vision Encoders.} We note that two BLIP-L vision encoders are used and compared to other baselines. During the pre-training stage, our model is compared with the BLIP-L base, which processes images at a size of $224 \times 224$ pixels. In the fine-tuning stage, since other approaches use an enhanced BLIP-L, we also train another CoLLM variant using the BLIP-L fine-tuned on COCO~\cite{lin2014microsoft} captions. This variant utilizes a larger image size of $384 \times 384$ pixels.

The performance differences between these variants are presented in~\Tref{tab:blip}. A significant gap can be observed between the two variants, particularly in the CIRCO metrics.

\begin{table}[t!]
    \centering
    \caption{Performance of different BLIP-L vision encoders after pre-training with image-caption pairs. The model demonstrates a significant improvement when utilizing a more advanced image encoder.}
    \footnotesize
    \begin{tblr}{width=\linewidth,colspec={@{}X[4,l]|X[1,c]X[1,c]X[1,c]|X[1,c]X[1,c]X[1,c]|X[1,c]X[1,c]@{}},stretch = 0,colsep=2pt}
    \toprule
         \SetCell[r=2]{l} BLIP-L Variants
         & \SetCell[c=3]{c} CIRCO (mAP$\uparrow$) & &
            & \SetCell[c=3]{c} CIRR (Rec.$\uparrow$) & &
                & \SetCell[c=2]{c} FIQ (Rec.$\uparrow$) \\
    \hline
    & @5 & @10 & @50 &@1 &@10 & @50 & @10 & @50 \\
    \hline
    Base & 19.4 & 20.4 & 23.3 & 35.1 & 78.6 & 94.2 & 34.6 & 56.0\\
    Fine-tuned COCO & 26.0 & 26.7 & 29.9 & 41.8 & 81.9 & 95.3 & 37.0 & 57.4 \\
    \bottomrule
    \end{tblr}
    \label{tab:blip}
\end{table}

\mypara{Additional Quantitative Results.} In~\Tref{tab:pretrain_fiq}, we provide detailed results of models evaluated on the Fashion-IQ Validation set without training on CIR triplet datasets. This table extends~\Tref{tab:pretrain}. Our models achieve the best results on most metrics when using CLIP-L and BLIP-L vision encoders. For CLIP-B, our CoLLM ranks second in the dress and shirt categories.

We also examine the effect of LoRA-tuning on different vision encoders during pre-training, as shown in~\Tref{tab:lora_setting}. While CLIP models show significant improvement with vision encoder tuning, BLIP-L exhibits a performance drop in both CIRCO and Fashion-IQ. This issue may stem from BLIP's synthetic captions. CLIP models, trained on noisier captions, benefit from further tuning. In contrast, BLIP, as a more advanced model, is already well-trained, and additional vision encoder tuning on a smaller dataset might lead to overfitting.

\begin{table}[t!]
    \centering
    \caption{Full results of Fashion-IQ validation, extension of~\Tref{tab:pretrain}. \textbf{Bold} and \underline{underlined} values indicate the best and second-best scores within each vision encoder group. Models that incorporate LLMs in their architectures are marked with $^\star$, and results reproduced by our team are denoted with $^\ddagger$.}
    \footnotesize
    \begin{tblr}{width=\linewidth,colspec={@{}X[3,l]|X[1,c]X[1,c]|X[1,c]X[1,c]|X[1,c]X[1,c]@{}},stretch = 0,colsep=2pt,row{3,9,17,20}={myorange},row{8,16,19, 21}={mypink}}
    \toprule
         \SetCell[r=2]{l} Model
         & \SetCell[c=2]{c} Dress & 
            & \SetCell[c=2]{c} Shirt &
                & \SetCell[c=2]{c} Toptee \\
    \hline
    & @10 & @50 & @10 & @50 & @10 & @50 \\
    \hline
    \SetCell[c=7]{c} OpenAI CLIP-B/32  \\
    PALAVRA~\cite{palavra} & 17.3 & 35.9 & 21.5 & 37.1 & 20.6 & 38.8  \\ 
    SEARLE~\cite{searle} & 18.5 & 39.5 & 24.4 & 41.6 & 25.7 & 46.5  \\
    Slerp-TAT~\cite{slerptat} & 19.2 & 42.1 & 23.1 & 42.0 & \snd{26.6} & \snd{47.8}  \\ 
    CIReVL$^\star$~\cite{cirevl} & \fst{25.3} & \fst{46.4} & \fst{28.4} & \fst{47.8} & \fst{31.2} & \fst{53.9}  \\
    CoLLM$^\star$ & \snd{22.9} & \snd{43.8} & \snd{24.9} & \snd{45.1} & 26.4 & 46.8 \\
    \hline
    \SetCell[c=7]{c} OpenAI CLIP-L/14  \\
    Pic2World~\cite{pic2word} & 20.0 & 40.2 & 26.2 & 43.6 & 27.9 & 47.4  \\ 
    SEARLE~\cite{searle} & 20.5 & 43.1 & 26.9 & 45.6 & 29.3 & 50.0  \\ 
    LinCIR$^\ddagger$~\cite{lincir} & 20.9 & 41.9 & 29.2 & 47.4 & 29.2 & 50.5 \\ 
    ContextI2W~\cite{contexti2w} & 23.1 & \snd{45.3} & \snd{29.7} & 48.6 & 30.6 & \snd{52.9}  \\ 
    Slerp-TAT~\cite{slerptat} & 23.4 & 45.1 & 29.6 & 46.5 & \snd{32.0} & 51.2  \\ 
    CIReVL$^\star$~\cite{cirevl} & \fst{24.8} & 44.8 & 29.5 & \snd{47.4} & 31.4 & \fst{53.7}  \\
    CoLLM$^\star$ & \snd{24.6} & \fst{46.5} & \fst{33.4} & \fst{50.5} & \fst{32.4} & 51.6 \\
    \hline
    \SetCell[c=7]{c} BLIP-L/16  \\
    Slerp-TAT~\cite{slerptat} & \snd{29.2} & \snd{50.6} & \snd{32.1} & \snd{51.6} & \snd{37.0} & \snd{57.7} \\
    CoLLM$^\star$ & \fst{30.8} & \fst{53.8} & \fst{34.2} & \fst{53.9} & \fst{38.7} & \fst{60.2} \\
    \hline
    \SetCell[c=7]{c} BLIP-L/16 $384\x 384$; fine-tuned COCO  \\
    CoLLM$^\star$ & 32.7 & 54.1 & 38.1 & 57.5 & 40.3 & 61.0 \\
    \bottomrule
    \end{tblr}
    \label{tab:pretrain_fiq}
\end{table}

\begin{table*}[t!]
    \centering
    \caption{Performance of CoLLM with different vision encoder and LoRA tuning applied to Vision Encoder (ViT). CLIP models require ViT tuning to achieve optimal performance, whereas BLIP-L performs better with a frozen ViT. \textbf{Bold} denotes the best score for each vision encoder.}
    \footnotesize
    \begin{tblr}{width=\linewidth,colspec={@{}X[3,l]X[1,c]|X[1,c]X[1,c]X[1,c]|X[1,c]X[1,c]X[1,c]|X[1,c]X[1,c]|X[1,c]X[1,c]|X[1,c]X[1,c]|X[1,c]X[1,c]@{}},stretch = 0,colsep=2pt}
    \toprule
    \SetCell[r=3]{l} Vision Encoder 
        & \SetCell[r=3]{c} LoRA ViT
            & \SetCell[r=2,c=3]{c} CIRCO (mAP$\uparrow$) & &
                & \SetCell[r=2,c=3]{c} CIRR (Recall$\uparrow$) & &
                & \SetCell[c=8]{c} Fashion-IQ (Recall$\uparrow$) \\
    \hline
        & & & & & & & &
        \SetCell[c=2]{c} Dress & 
            & \SetCell[c=2]{c} Shirt &
                & \SetCell[c=2]{c} Toptee & 
                  & \SetCell[c=2]{c} Average \\
    \hline
    & & @5 & @10 & @50 & @1 & @10 & @50 & @10 & @50 & @10 & @50 & @10 & @50 & @10 & @50 \\
    \hline
    OpenAI CLIP-B/32 & & 11.7 & 12.0 & 13.7 
                        & 23.2 & 67.4 & 91.1 
                        & 20.3 & 40.2 & 23.8 & 42.1 & 24.7 & 42.6 & 22.9 & 41.6 \\
    OpenAI CLIP-B/32 & \checkmark & \fst{12.9} & \fst{13.2} & \fst{15.0}
                        & \fst{28.6} & \fst{71.8} & \fst{92.7}
                        & \fst{22.9} & \fst{43.8} & \fst{24.9} & \fst{45.1} & \fst{26.4} & \fst{46.8} & \fst{24.8} & \fst{45.2} \\
    \hline
    OpenAI CLIP-L/14 & & 16.1 & 16.9 & 19.1 
                        & 24.5 & 69.2 & 90.9 
                        & 23.5 & 42.4 & 32.7 & 49.1 & 29.8 & 48.9 & 28.7 & 46.8 \\
    OpenAI CLIP-L/14 & \checkmark & \fst{20.3} & \fst{20.8} & \fst{23.4}
                        & \fst{29.7} & \fst{72.8} & \fst{91.5}
                        & \fst{24.6} & \fst{46.5} & \fst{33.4} & \fst{50.5} & \fst{32.4} & \fst{51.6} & \fst{30.1} & \fst{49.5} \\
    \hline
    BLIP-L/16 & & \fst{19.4} & \fst{20.4} & \fst{23.3} 
                & 35.1 & 78.6 & 94.2 
                & \fst{30.8} & \fst{53.8} & 34.2 & 53.9 & \fst{38.7} & \fst{60.2} & \fst{34.6} & \fst{56.0} \\
    BLIP-L/16 & \checkmark & 18.6 & 19.4 & 22.1 
                    & \fst{37.7} & \fst{79.2} & \fst{94.6} 
                    & 30.6 & 53.4 & \fst{34.4} & \fst{54.1} & 37.3 & 59.7 & 34.1 & 55.7 \\
    \bottomrule
    \end{tblr}
    \label{tab:lora_setting}
\end{table*}

\subsection{Fine-tuning}
\mypara{Implementation Details.} To ensure a fair comparison across models and datasets, we implement several adjustments in our training process. We reduce the number of trainable parameters by setting the LoRA rank and alpha to 16. At this stage, only the LLM is fine-tuned, as the vision encoder features are already aligned during the pre-training phase. Other settings remain consistent with the pre-training stage. For the BLIP-L vision encoder, we use BLIP-L/16 fine-tuned on COCO captions and increase the image input size to $384 \times 384$ pixels, aligning with prior methodologies.

For experiments involving LaSCo and our MTCIR, both BLIP-L and CoLLM models are trained for one epoch. We utilize the publicly available BLIP-L weights pretrained on WebCoVR. Despite an imbalance in sample size between WebCoVR and LaSCo, extending the training beyond one epoch for LaSCo is impractical, as the model rapidly overfits after the initial epoch (see \Tref{tab:lasco_overfit}).
\begin{table}[t!]
    \centering
    \caption{BLIP-L/16 ($384 \times 384$) fine-tuned on COCO exhibits rapid overfitting on the LaSCo dataset after the first training epoch. Performance is measured by Recall Sum on CIRR validation set (@1,10,50) and Fashion-IQ (@10,50).}
    \footnotesize
    \begin{tblr}{
        colspec={@{}X[2,l]*{5}{X[1,c]}@{}},
    }
    \toprule
    Epoch & 1 & 2 & 3 & 4 & 5 \\
    \hline
    CIRR Val & \textbf{216.2} & 214.3 & 214.8 & 212.7 & 213.4 \\
    Fashion-IQ & \textbf{68.9} & 63.9 & 62.7 & 62.5 & 62.6 \\
    \bottomrule
    \end{tblr}
    \label{tab:lasco_overfit}
\end{table}

\mypara{Additional Quantitative Results.} We provide details of models fine-tuned on synthetic CIR datasets in~\Tref{tab:ft_fiq}. Our CoLLM with the BLIP-L vision encoder achieves the best overall performance across most metrics, even surpassing models equipped with larger vision encoders. Using CLIP-L vision encoders, our model achieves the best scores in half of the metrics compared to other methods.

Fashion-IQ detailed results from~\Tref{tab:ft_data} are also presented in~\Tref{tab:ft_data_fiq}. Our MTCIR consistently enhances the performance of both models across all sub-category metrics of Fashion-IQ. For completeness, we also report the models' performance on the CIRCO benchmark in~\Tref{tab:ft_circo}. However, we note that CIRCO is not an ideal benchmark for these models due to data leakage concerns. Despite this, our models achieve strong performance, even though other works may have been trained on a subset of the CIRCO images.

\Tref{tab:ft_pretrain} illustrates the performance drop when the BLIP-L/16 model with resolution $384 \x 384$, initially finetuned on COCO, is directly trained on the MTCIR dataset without further pretraining. While the model trained solely on MTCIR still surpasses previous works shown in~\Tref{tab:ft}, incorporating a pretraining stage results in substantial improvements in performance metrics.

\begin{table}[t!]
    \centering
    \caption{Full results of Fashion-IQ validation, extension of~\Tref{tab:ft}. \textbf{Bold} is used to highlight the best overall scores, while \underline{underline} marks the best metrics within the same vision encoder group.}
    \footnotesize
    \begin{tblr}{width=\linewidth,colspec={@{}X[3,l]|X[1,c]X[1,c]|X[1,c]X[1,c]|X[1,c]X[1,c]@{}},stretch = 0,colsep=2pt,row{3,5,7,11}={myorange},row{10,13}={mypink}}
    \toprule
         \SetCell[r=2]{l} Model
         & \SetCell[c=2]{c} Dress & 
            & \SetCell[c=2]{c} Shirt &
                & \SetCell[c=2]{c} Toptee \\
    \hline
    & @10 & @50 & @10 & @50 & @10 & @50 \\
    \hline
    \SetCell[c=7]{c} CoCA-L/18 $288\x 288$ \\
    MagicLens~\cite{magiclens} & 32.3 & 52.7 & 40.5 & 59.2 & 41.4 & 63.0 \\
    \hline
    \SetCell[c=7]{c} EVA-CLIP ViT-G/14 $364\x 364$ \\
    CoVR2~\cite{ventura24covr2} & 34.3 & 56.2 & \fst{41.2} & 59.3 & 39.0 & 59.8 \\
    \hline
    \SetCell[c=7]{c} OpenAI CLIP-L/14 $224\x 224$ \\
    CompoDiff~\cite{compodiff} & 32.2 & 46.3 & \snd{37.7} & 49.1 & \snd{38.1} & 50.6 \\
    MagicLens~\cite{magiclens} & 25.5 & 46.1 & 32.7 & 53.8 & 34.0 & \snd{57.7} \\
    CoLLM & \snd{28.1} & \snd{51.6} & 36.3 & \snd{55.8} & 34.4 & 55.1 \\
    \hline
    \SetCell[c=7]{c} BLIP-L/16 $384\x 384$; fine-tuned on COCO \\
    Omkar et al.~\cite{thawakar2024composed} & 24.6 & 40.9 & 33.1 & 48.4 & 33.2 & 50.2 \\
    CoLLM & \fst{\snd{35.8}} & \fst{\snd{58.9}} & \snd{39.6} & \fst{\snd{59.5}} & \fst{\snd{42.0}} & \fst{\snd{63.8}} \\
    \bottomrule
    \end{tblr}
    \label{tab:ft_fiq}
\end{table}

\begin{table}[t!]
    \centering
    \caption{Performance of models training on synthetic datasets on CIRCO benchmark.}
    \footnotesize
    \begin{tblr}{width=\linewidth,colsep=4pt,colspec={@{}X[3,l]|X[4,l]|X[1,c]X[1,c]X[1,c]},row{10,14}={mypink},row{3,5,7,11}={myorange},stretch = 0}
    \toprule
         \SetCell[r=2]{l} Method
         & \SetCell[r=2]{l}{Dataset}
            & \SetCell[c=3]{c} CIRCO  (mAP$\uparrow$) \\
    \hline
    & & @5 & @10 & @50 \\
    \hline
    \SetCell[c=5]{c} CoCa-L/18 $288\x 288$ \\
    MagicLens~\cite{magiclens} & MagicLens~\cite{magiclens} & \fst{34.1} & \fst{35.4} & \fst{39.2} \\ 
    \hline
    \SetCell[c=5]{c} EVA-CLIP ViT-G/14 $364\x364$ \\
     CoVR2~\cite{ventura24covr2} & WV-CC-CoVIR~\cite{ventura24covr2} & 28.3 & 29.6 & 33.3  \\
    \hline
    \SetCell[c=5]{c} OpenAI CLIP-L/14 $224\x 224$ \\
    CompoDiff~\cite{compodiff} & 
    SynTrip18M~\cite{compodiff} & 12.6 & 13.4 & 16.4 \\ 
    MagicLens~\cite{magiclens} &  MagicLens~\cite{magiclens}  & \snd{29.6} & \snd{30.8} & \snd{34.4} \\ 
    CoLLM & MTCIR (ours) & 24.4 & 25.2 & 28.2 \\
    \hline
    \SetCell[c=5]{c} BLIP-L/16 $384\x 384$; fine-tuned on COCO \\
    CoVR~\cite{webcovr} & WebCoVR~\cite{webcovr} & 21.4 & 22.3 & 25.5 \\
    CoLLM & MTCIR (ours) & \snd{29.0} & \snd{29.8} & \snd{33.4} \\ 
    \bottomrule
    \end{tblr}
    \label{tab:ft_circo}
\end{table}

\begin{table}[t!]
    \centering
    \caption{Full results of Fashion-IQ validation, extension of~\Tref{tab:ft_data}. \textbf{Bold} values indicate the best score within each method group.}
    \footnotesize
    \begin{tblr}{width=\linewidth,colspec={@{}X[3,l]|X[1,c]X[1,c]|X[1,c]X[1,c]|X[1,c]X[1,c]@{}},stretch = 0,colsep=2pt,row{3,7}={myorange},row{6,9}={mypink}}
    \toprule
         \SetCell[r=2]{l} Dataset
         & \SetCell[c=2]{c} Dress & 
            & \SetCell[c=2]{c} Shirt &
                & \SetCell[c=2]{c} Toptee \\
    \hline
    & @10 & @50 & @10 & @50 & @10 & @50 \\
    \hline
    \SetCell[c=7]{c} BLIP-L~\cite{blip} \\
    LaSCo~\cite{caselasco} & 20.2 & 38.5 & 26.3 & 43.3 & 28.0 & 50.3 \\
    WebCoVR~\cite{webcovr} & 22.0 & 39.1 & 30.5 & 46.1 & 27.7 & 44.7 \\
    MTCIR (ours) & \fst{32.3} & \fst{55.3} & \fst{40.6} & \fst{58.8} & \fst{40.9} & \fst{63.4} \\
    \hline
    \SetCell[c=7]{c} CoLLM \\
    LaSCo~\cite{caselasco} & 34.9 & 58.2 & 38.8 & 58.8 & 41.8 & 63.4 \\
    MTCIR (ours) & \fst{35.8} & \fst{58.9} & \fst{39.6} & \fst{59.5} & \fst{42.0} & \fst{63.8} \\
    \bottomrule
    \end{tblr}
    \label{tab:ft_data_fiq}
\end{table}

\begin{table}[h!]
    \centering
    \caption{Performance of CoLLM (with BLIP-L/16 $384\x 384$ finetuned on COCO) is superior when pre-training on 5M image-caption pairs.}
    \footnotesize
    \begin{tblr}{width=\linewidth,colspec={@{}c|ccc|cc|ccc|cc@{}},stretch = 0,colsep=2.5pt}
    \toprule
    \SetCell[r=2]{c} Pre-train
    & \SetCell[c=3]{c} CIRR Test & &
    & \SetCell[c=2]{c} FIQ &
    & \SetCell[c=3]{c} Ref. CIRR & &
    & \SetCell[c=2]{c} Ref. FIQ & \\
    \hline
    &@1 &@10 & @50 & @10 & @50 & @1 &@10 & @50 & @10 & @50 \\
    \hline
    Yes & 45.8 & 84.7 & 95.9 & 39.1 & 60.7 & 60.7 & 92.7 & 98.2 & 57.2 & 76.4 \\
    No & 42.0 & 81.8 & 95.6 & 34.7 & 56.3 & 55.4 & 90.7 & 97.8 & 52.1 & 73.4 \\
    \bottomrule
    \end{tblr}
    \label{tab:ft_pretrain}
    \vspace{-1.2em}
\end{table}

\begin{figure*}[t]
    \centering
    \includegraphics[width=0.75\linewidth]{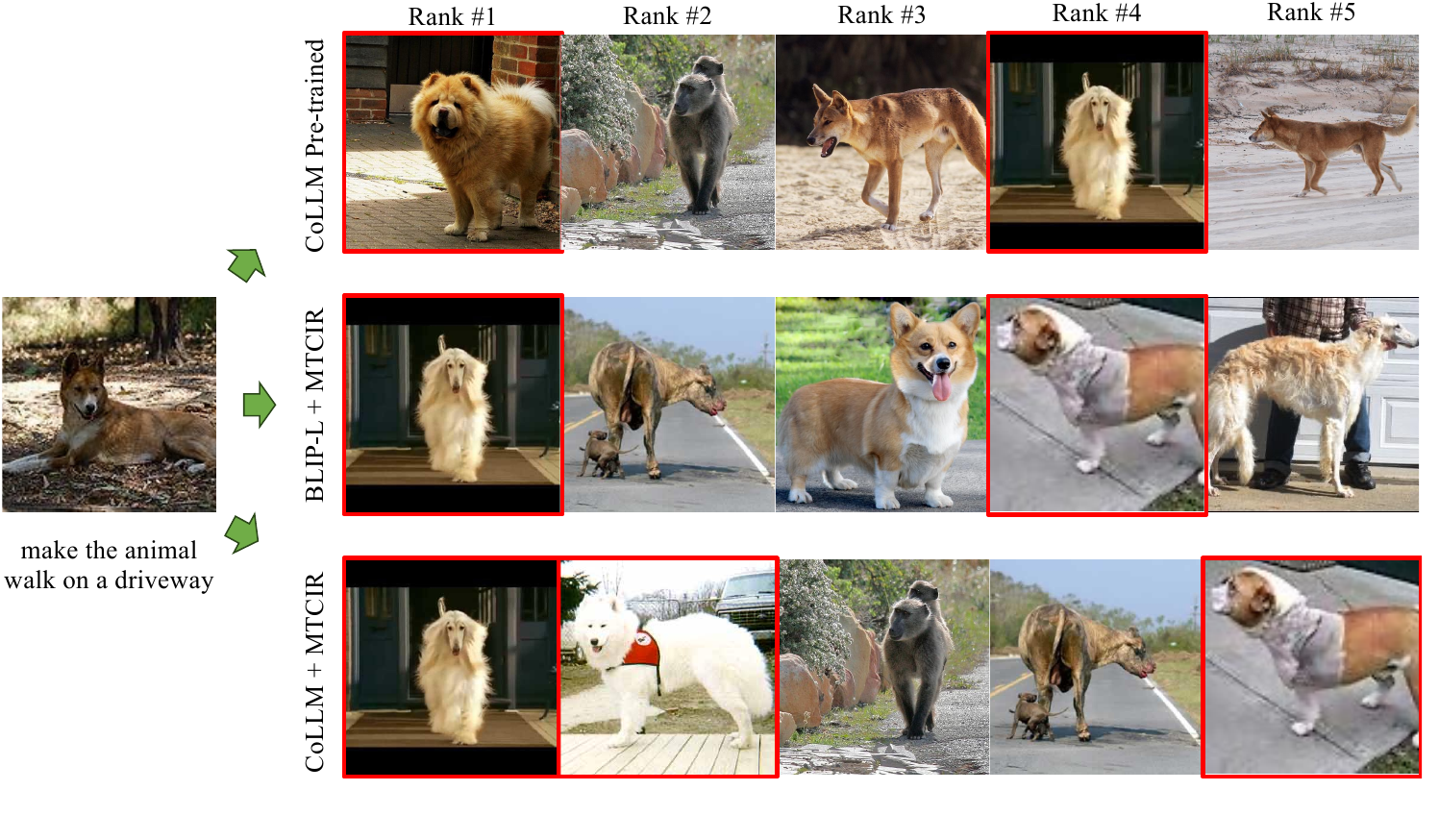}
    \includegraphics[width=0.75\linewidth]{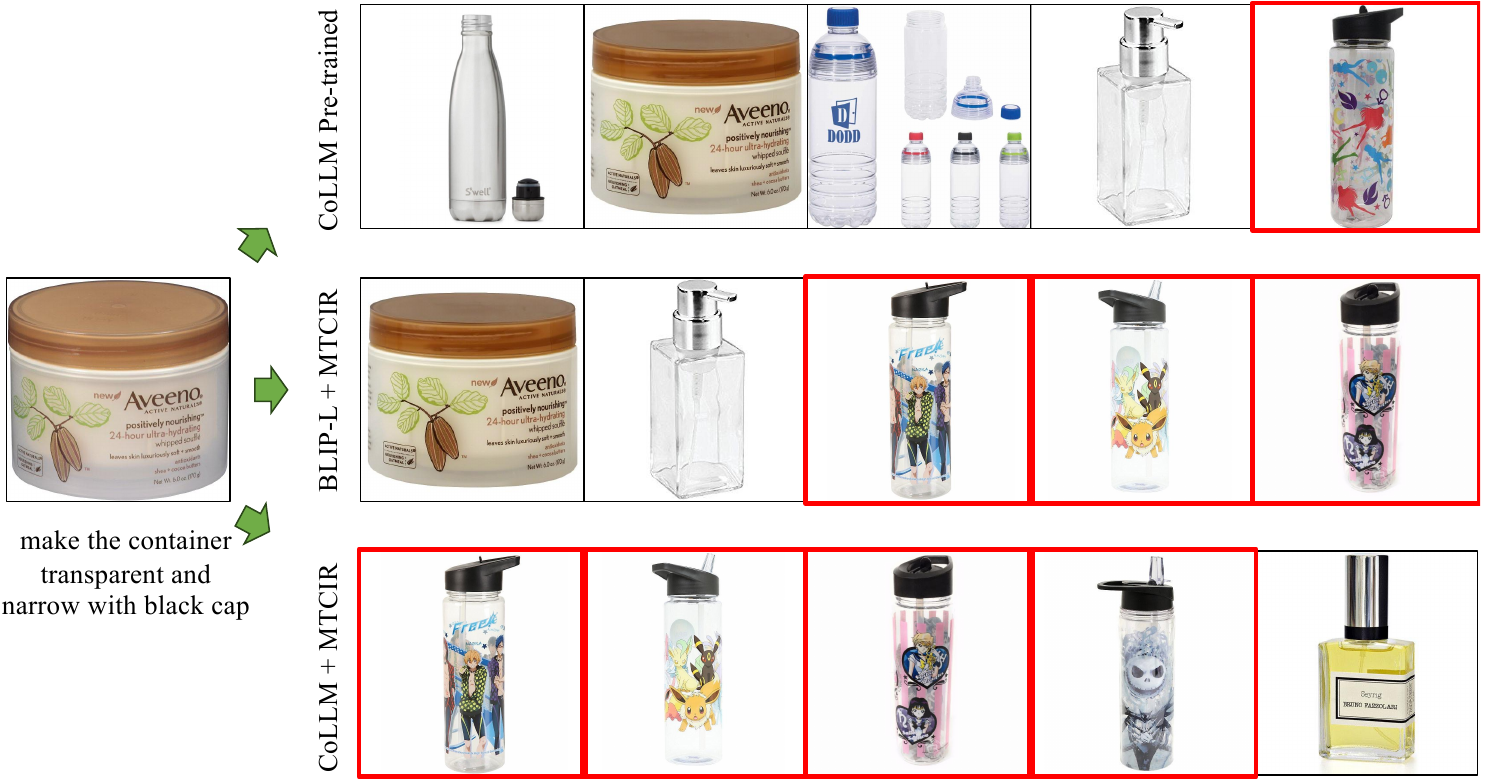}
    \caption{Retrieval results of Pre-trained CoLLM,BLIP-L fine-tuned on MTCIR (BLIP-L + MTCIR) and CoLLM fine-tuned on MTCIR (CoLLM + MTCIR) on CIRR Test set. \textcolor{red}{Red} highlights potential correct results (since we do not have the ground-truth on that test set). }
    \label{fig:performance_cirr}
\end{figure*}

\begin{figure*}[t]
    \centering
    \includegraphics[width=0.75\linewidth]{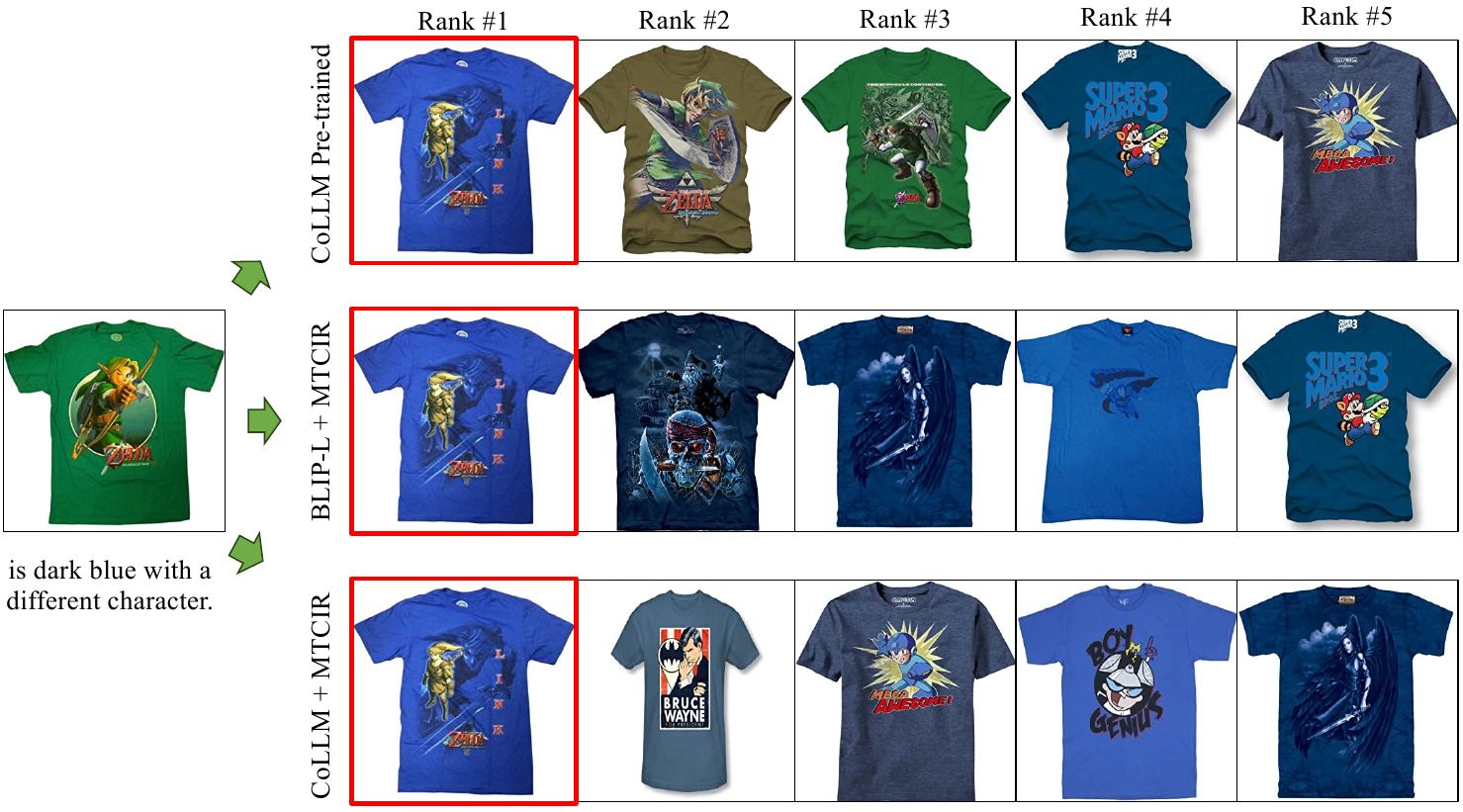}
    \includegraphics[width=0.75\linewidth]{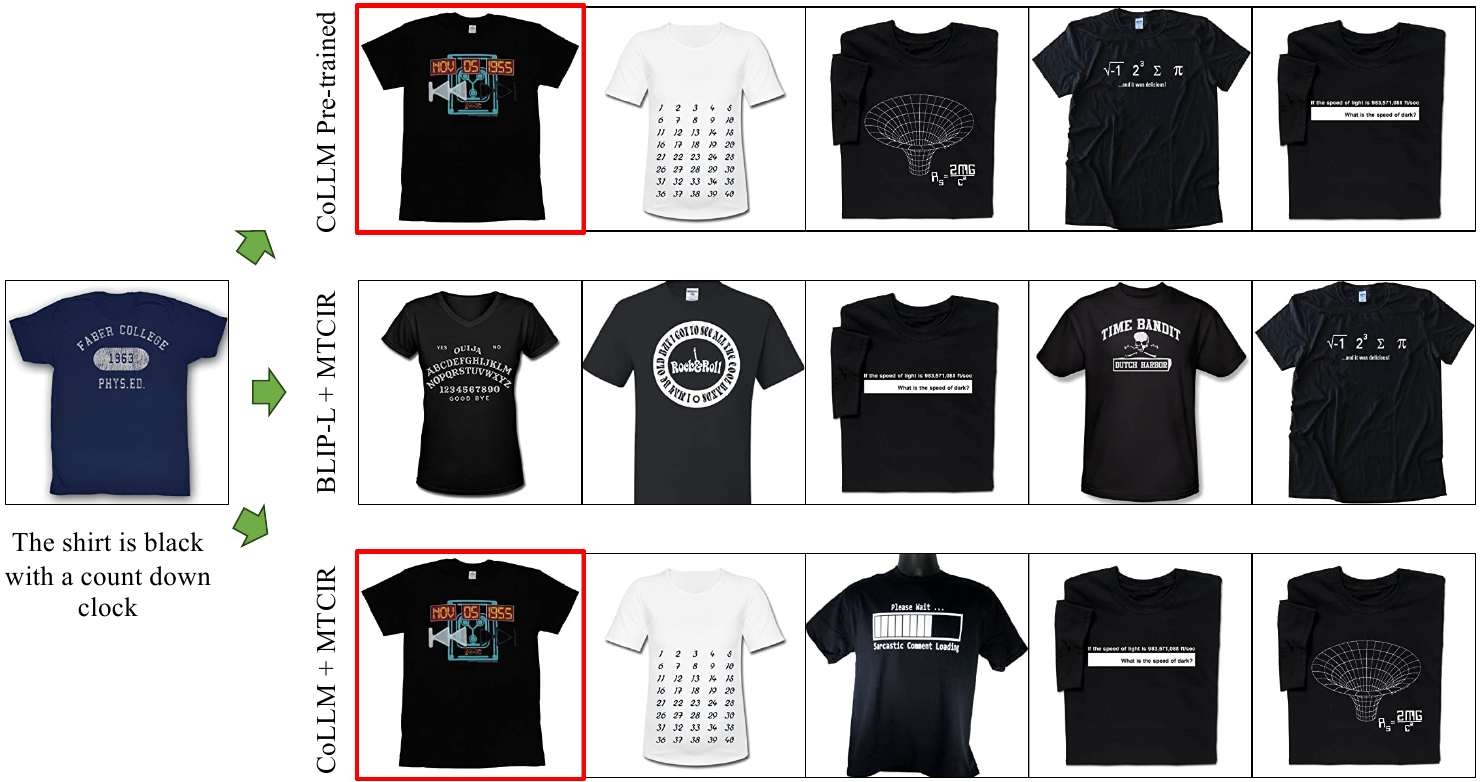}
    \caption{Retrieval results of Pre-trained CoLLM, BLIP-L fine-tuned on MTCIR (BLIP-L + MTCIR) and CoLLM fine-tuned on MTCIR (CoLLM + MTCIR) on Fashion-IQ Validation set. \textcolor{red}{Red} highlights the ground-truth.}
    \label{fig:performance_fiq}
\end{figure*}

\mypara{Qualitative Results.} \Fref{fig:performance_cirr} and \Fref{fig:performance_fiq} present a performance comparison of CoLLM after the pre-training stage, BLIP-L, and our CoLLM fine-tuned on the respective datasets. All models use the BLIP-L/16-384 vision encoder fine-tuned on COCO.

The pre-trained model already demonstrates reasonable performance, while the fine-tuned version retrieves a higher number of correct images. Although BLIP-L achieves good results, it struggles with capturing precise image details in some cases (e.g., the second samples in \Fref{fig:performance_cirr} and \Fref{fig:performance_fiq}).

\subsection{Refined benchmarks}

\begin{figure}[t]
    \centering
    \includegraphics[width=\linewidth]{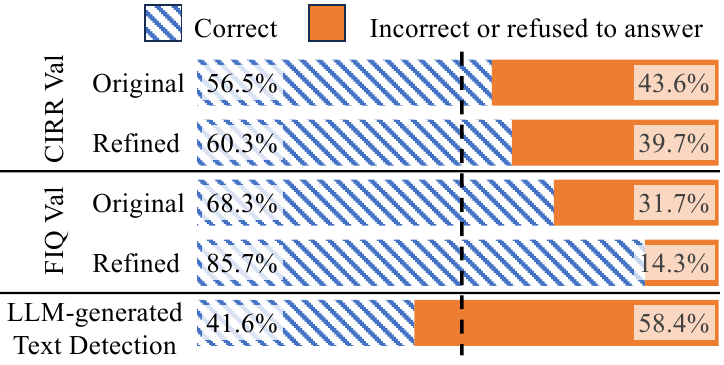}
    \caption{Human studies in evaluating refined benchmarks. Our new refined benchmarks increase the number of correct answers while human finds difficulty in detect AI-generated text.}
    \label{fig:user_study}
\end{figure}

\mypara{Human Studies on Quality.} As detailed in~\Sref{sec:new_benchmarks}, we have improved the CIRR~\cite{cirr} and Fashion-IQ~\cite{fashioniq} validation benchmarks. To evaluate the quality of the newly generated texts in the refined benchmarks, we conducted human studies on random samples from the Regenerated group. The results are summarized in~\Fref{fig:user_study}:

\begin{enumerate}
    \item \textit{Refined CIRR Evaluation:} We used the same strategy as the validation step (Step 1) during the benchmark refinement process. Seven random regenerated samples, along with their original texts, were selected. Participants were asked to identify the target image using the reference image and either the regenerated or original modification text. Alongside the target image, two of its most similar images were included as options. Participants could refuse to answer if they believed there was no or more than one correct answer. From 130 responses, the new refined CIRR benchmark reduced ambiguity, increasing the correct answers by 4\%.
    \item \textit{Refined Fashion-IQ Evaluation:} A similar process was used for the Fashion-IQ dataset, with 12 questions (4 per category). From 130 collected responses, the refined benchmark significantly addressed the issues in the original dataset, increasing the proportion of correct answers by approximately 17\%.
    \item \textit{LLM-generated Text Quality Evaluation:} Participants were tasked with identifying which text was more likely generated by LLMs from nine pairs of old and new modification texts across both CIRR and Fashion-IQ datasets. Participants could also refuse to answer. From 125 responses, over half either selected incorrect answers or were unable to distinguish LLM-generated texts, as shown in the last row of~\Fref{fig:user_study}.
\end{enumerate}

These surveys validate our assumptions in improving benchmarks. Firstly, the existing evaluation sets contain ambiguities that even humans struggle to resolve. Secondly, regenerating texts significantly reduces these ambiguities, as evidenced by the improvement in human accuracy. Lastly, the newly generated texts align closely with human language, as demonstrated by participants' difficulty in identifying AI-generated texts. This highlights the effectiveness of our pipeline in creating refined and natural benchmarks.

\mypara{Benchmark Ambiguity.}
\Tref{tab:abl_query_type} presents the performance discrepancy across different evaluation queries on both the original and refined benchmarks, following the analysis in~\cite{caselasco}. The BLIP-L/16 ($384 \times 384$) model, fine-tuned on COCO, is evaluated after training on the MTCIR dataset. Notably, using only the modification text in the query yields high performance in both benchmarks. One possible explanation is that paired images share fewer common features, making the text a crucial factor in retrieval. This observation highlights a potential research direction for improving benchmark design.
\begin{table}[t!]
    \centering
    \caption{Performance of different query types on the original and refined benchmarks of BLIP-L/16 $384\x 384$ fine-tuned on COCO.}
    \footnotesize
    \begin{tblr}{width=\linewidth,colspec={@{}c|ccc|cc|ccc|cc@{}},stretch = 0,colsep=2.5pt}
    \toprule
    \SetCell[r=2]{c} Query
    & \SetCell[c=3]{c} CIRR Val & &
    & \SetCell[c=2]{c} FIQ &
    & \SetCell[c=3]{c} Ref. CIRR & &
    & \SetCell[c=2]{c} Ref. FIQ & \\
    \hline
    &@1 &@10 & @50 & @10 & @50 & @1 &@10 & @50 & @10 & @50 \\
    \hline
    Composed & 43.8 & 84.1 & 95.4 & 37.9 & 59.2 & 58.0 & 91.6 & 97.9 & 56.8 & 76.6 \\
    Text &  38.5 & 75.1 & 89.3 & 28.4 & 48.5 & 52.8 & 86.7 & 95.0 & 49.9 & 69.9 \\
    \bottomrule
    \end{tblr}
    \label{tab:abl_query_type}
\end{table}

\mypara{Additional Quantitative Results.} \Tref{tab:reeval_fiq} presents the recall metrics for all Fashion-IQ categories, extending~\Tref{tab:reeval} from the main paper. Our MTCIR continues to enhance model performance, achieving the best results across most metrics. Notably, CoLLM fine-tuned on MTCIR achieves the best overall results, outperforming both CoVR2 and MagicLens, despite utilizing a smaller fine-tuned dataset.

\begin{table}[t!]
    \centering
    \caption{Performance of models on all categories of refined Fashion-IQ validation set. This is an extension of~\Tref{tab:reeval}. \textbf{Bold} indicates the highest score, while \underline{underlined} values represent the best metric within the same vision encoder group. }
    \footnotesize
    \begin{tblr}{width=\linewidth,colsep=3pt,colspec={@{}X[4,l]|X[4.5,l]|X[1,c]X[1,c]|X[1,c]X[1,c]|X[1,c]X[1,c]},row{7,11}={mypink},row{3,5,8}={myorange},stretch = 0}
    \toprule
         \SetCell[r=2]{l} Method
         & \SetCell[r=2]{l}{Dataset}
            & \SetCell[c=2]{c} Dress &
                & \SetCell[c=2]{c} Shirt & 
                    & \SetCell[c=2]{c} Toptee \\
    \hline
    & & @10 & @50 & @10 & @50 & @10 & @50 \\
    \hline
    \SetCell[c=8]{c} EVA-CLIP ViT-G/14 $364\x364$ \\
     CoVR2~\cite{ventura24covr2} & WV-CC-VIR~\cite{ventura24covr2} & 48.6 & 69.8 & \fst{58.5} & 74.7 & 55.4 & 74.2 \\
    \hline
    \SetCell[c=8]{c} OpenAI CLIP-L/16 $224\x 224$ \\
    MagicLens~\cite{magiclens} &  MagicLens~\cite{magiclens}  &  38.0 & 62.6 & 49.1 & 69.9 & 49.5 & \fst{71.9} \\ 
    CoLLM & MTCIR~(ours) & \snd{40.9} & \snd{64.4} & \snd{53.2} & \snd{71.1} & \snd{50.8} & 70.3 \\
    \hline
    \SetCell[c=8]{c} BLIP-L/16 $384\x 384$; fine-tuned on COCO \\
    BLIP-L~\cite{blip} & MTCIR~(ours)  & 48.1 & 70.6 & \snd{58.4} & 75.6 & 57.8 & 76.7 \\
    CoLLM & LaSCo~\cite{caselasco} & 52.2 & 72.9 & 57.6 & 75.1 & \fst{\snd{60.9}} & \fst{\snd{79.9}} \\ 
    CoLLM & MTCIR~(ours) & \fst{\snd{52.5}} & \fst{\snd{73.4}} & 58.2 & \fst{\snd{76.3}} & \fst{\snd{60.9}} & 79.4 \\ 
    \bottomrule
    \end{tblr}
    \label{tab:reeval_fiq}
\end{table}

\begin{figure*}[t]
    \centering
    \includegraphics[width=0.75\linewidth]{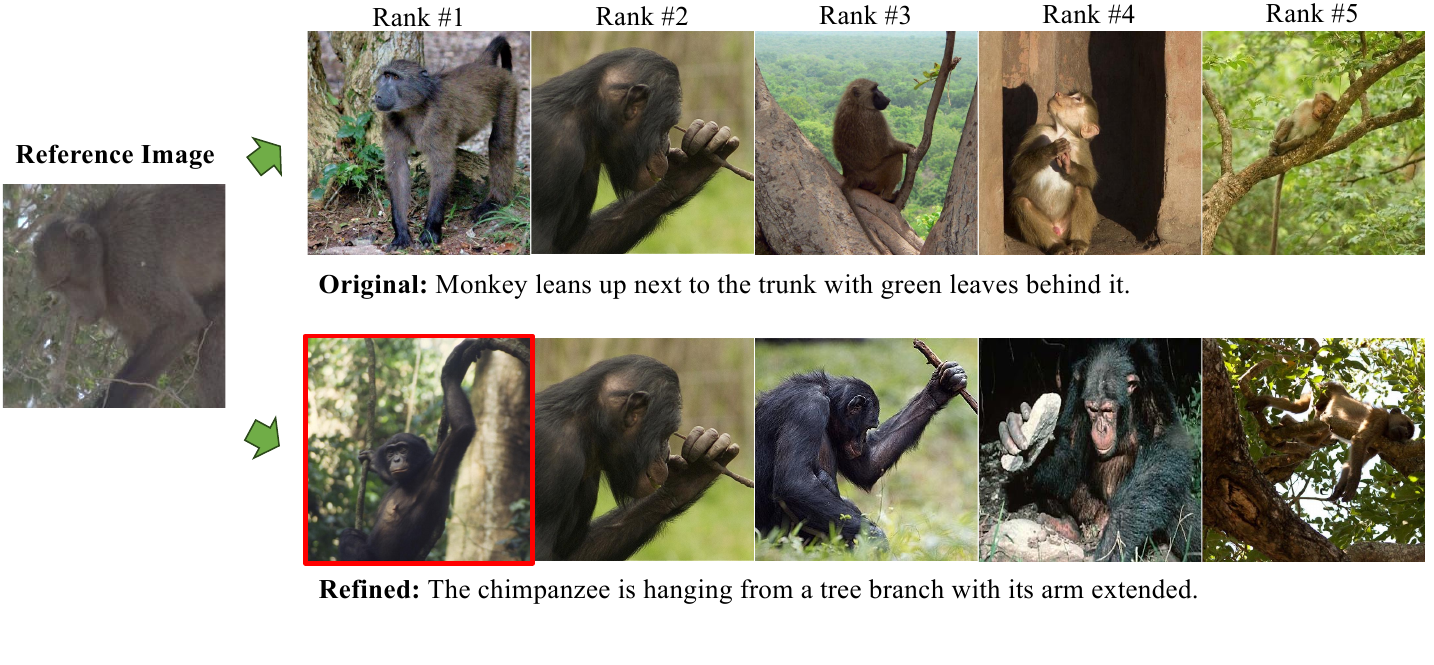}
    \includegraphics[width=0.75\linewidth]{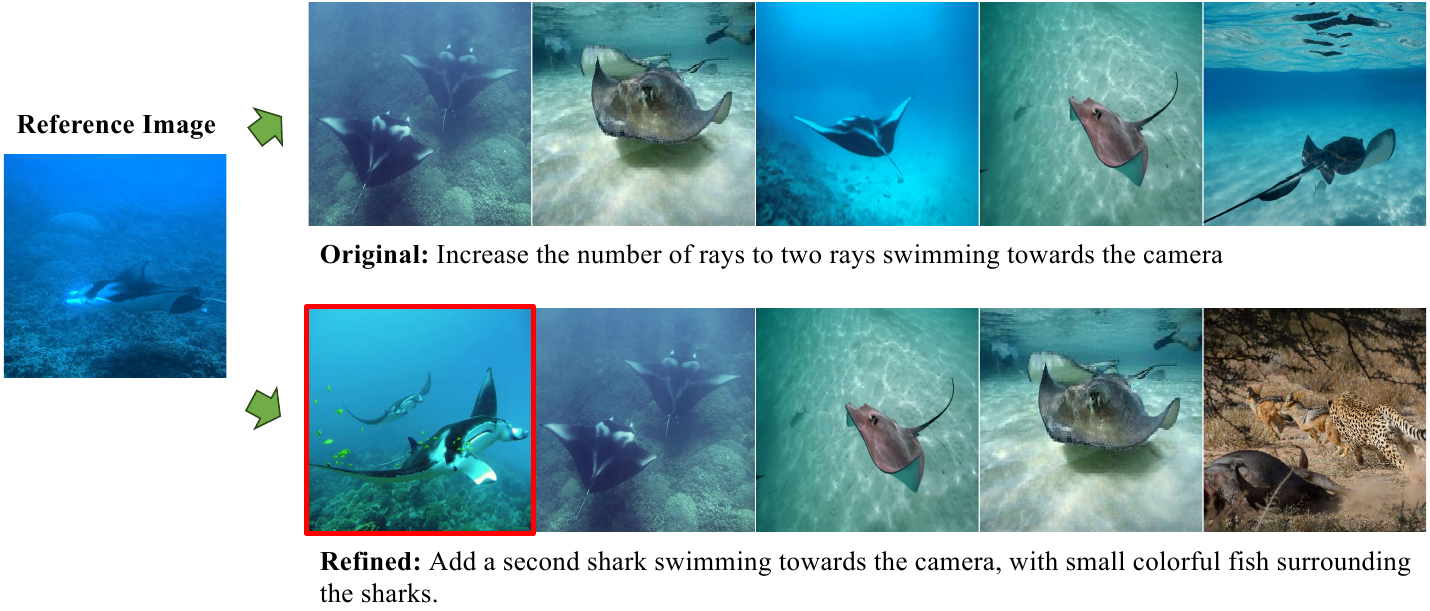}
    \caption{Retrieval results of CoLLM fine-tuned on MTCIR on original and Refined CIRR validation set. \textcolor{red}{Red} highlights the ground-truth. The new modification text helps the model to find the correct target images.}
    \label{fig:collm_refined_cirr}
\end{figure*}

\begin{figure*}[t]
    \centering
    \includegraphics[width=0.75\linewidth]{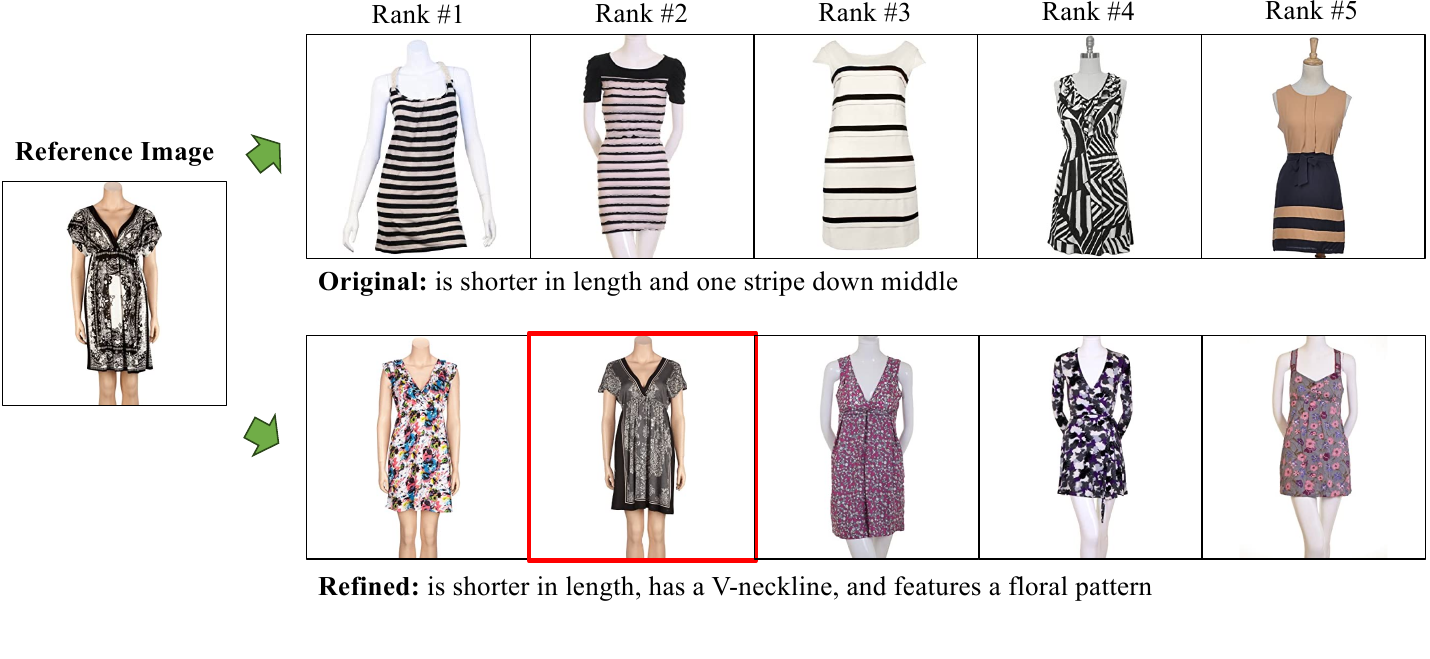}
    \includegraphics[width=0.75\linewidth]{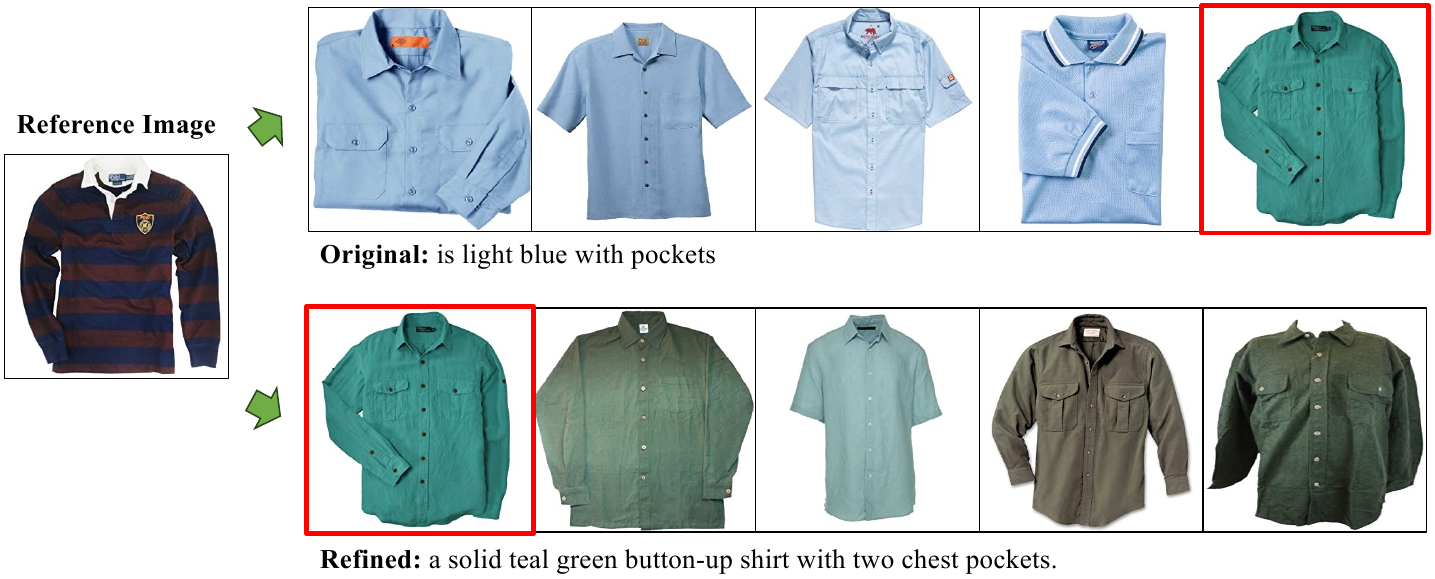}
    \caption{Retrieval results of CoLLM fine-tuned on MTCIR on original and Refined Fashion-IQ validation set. \textcolor{red}{Red} highlights the ground-truth. The new modification text with more details helps the model to find the correct target images.}
    \label{fig:collm_refined_fiq}
\end{figure*}

\mypara{Qualitative results.} The performance of CoLLM after fine-tuning with our MTCIR on the Refined CIRR and Fashion-IQ benchmarks is presented in~\Fref{fig:collm_refined_cirr} and~\Fref{fig:collm_refined_fiq}. The original modification texts are often ambiguous, lacking specific details needed to identify the correct target image. With refined modification texts, our model achieves superior results on both datasets. The new texts remain concise but provide more useful information, enabling the model to perform better.

\begin{table*}[t!]
    \centering
    \caption{Prompt structure to generate modification texts in MTCIR.}
    \footnotesize
    \begin{tblr}{width=\linewidth,colspec={@{}X[1,l]X[8,l]@{}},stretch = 0}
    \toprule
    System & \texttt{You are a language assistant that helps to generate the modification text between two image captions.} \\
    \hline
    Prompt & {
    \texttt{Generate the modified text for the following pair of image captions:} \\
    \texttt{Caption 1: [CAPTION 1]} \\
    \texttt{Caption 2: [CAPTION 2]} \\
    \texttt{<instruction>} \\
    \texttt{You need to answer in both forward, changes from image 1 to image 2, and backward, changes from image 2 to image 1, directions. The definition of each category and examples are as follows:} \\
    \texttt{1. [CATEGORY ID 1]: [CATEGORY DEFINITION 1]} \\
    \texttt{<example>} \\
    \texttt{Caption 1: [CAPTION EXAMPLE 1]}\\
    \texttt{Caption 2: [CAPTION EXAMPLE 2]}\\
    \texttt{Forward: [FORWARD EXAMPLE]}\\
    \texttt{Backward: [BACKWARD EXAMPLE]}\\
    \texttt{</example>} \\
    \texttt{...} \\
    \texttt{6. [CATEGORY ID 6]: [CATEGORY DEFINITION 6]} \\
    \texttt{...} \\
    \texttt{The text needs to be concise and details as you can see the images, not as you are reading the text. You should not add words "details, specific, description" to the text. Here are some bad examples:} \\
    \texttt{<example>} \\
    \texttt{[BAD EXAMPLES]} \\
    \texttt{</example>} \\
    \texttt{</instruction>} \\
    \texttt{One category can has multiple changes. For each change, you need to write one short sentence less than 20 words to describe the change. You need to answer all changes in the json format. Here is an example of the correct format:} \\
    \texttt{\{"forward": [\{"category": "number\_change", "text": "modified text"\},...],"backward": [\{"category": "number\_change","text": "modified text"\},...]\}}
    }
    \\ 
    \bottomrule
    \end{tblr}
    \label{tab:prompt}
\end{table*}

\begin{table*}[t!]
    \centering
    \caption{Prompt regenerating new modification texts for ambiguous samples in CIRR and Fashion-IQ.}
    \footnotesize
    \begin{tblr}{width=\linewidth,colspec={@{}X[1,l]X[8,l]@{}},stretch = 0}
    \toprule
    System & \texttt{You are the vision language bot that helps to generate the modification text given the reference image and the target image.} \\
    \hline
    Prompt & {
    \texttt{[REFERENCE IMAGE][TARGET IMAGE]} \\
    \texttt{You are given the reference image and the target image. The original modification text: "[MODIFICATION TEXT]" is bad and does not have enough details to find the target image. These are some examples of the modification text:} \\
    \texttt{<example>} \\
    \texttt{[GOOD SAMPLES]} \\
    \texttt{</example>} \\
    \texttt{Generate three new modification texts following the instruction below:}\\
    \texttt{<instruction>} \\
    \texttt{1. Understand the image content of the reference image (the first image).} \\
    \texttt{2. Understand the image content of the target image (the second image).} \\
    \texttt{3. text1: generate a short modification based on the original modification text with more specific details about the main information in the target image. It can be objects added or removed, colors, shapes or any other details.}\\
    \texttt{4. text2: add one more detail to the text1 without removing any information. It can be the information about the relationship between the objects in the target image, the background information.} \\
    \texttt{5. text3: add one more detail to the text2 without removing any information. It can be the view different from the reference image, any other details that is not in the first two texts.} \\
    \texttt{7. Answer in json format \{"text1": "new text 1", "text2": "new text 2", "text3": "new text 3"\}.} \\
    \texttt{</instruction>}\\
    \texttt{Again, note that the modification text should be short and concise.}
    } \\
    \bottomrule
    \end{tblr}
    \label{tab:benchmark_prompt_regen}
\end{table*}

\begin{table*}[t!]
    \centering
    \caption{Prompt validate sample ambiguity in CIRR and Fashion-IQ.}
    \footnotesize
    \begin{tblr}{width=\linewidth,colspec={@{}X[1,l]X[8,l]@{}},stretch = 0}
    \toprule
    System & \texttt{You are the vision language bot that helps to find the target image given reference image and modification text.} \\
    \hline
    Prompt & {
    \texttt{[REFERENCE IMAGE][CANDIDATE IMAGES]} \\
    \texttt{You are given the reference image and the candidate images. From the reference image and the modified text "[MODIFICATION TEXT]", find the best matched target image following the instruction below:} \\
    \texttt{<instruction>} \\
    \texttt{1. Understand the image content and the modification text.} \\
    \texttt{2. For each image in the candidate images, understand the image content.} \\
    \texttt{3. Find the best matched target image that matches the modification text.} \\
    \texttt{4. If there are two or more target images that are equally matched, answer -1.} \\
    \texttt{5. If the target image is not in the candidate images, answer -1.} \\
    \texttt{6. If the target image is in the candidate images, answer the index of the target image in the candidate images from left to right from 0 to 3.} \\
    \texttt{7. Answer in json format \{"answer": target image index, "explain": give the reason for each unmatched image\}.} \\
    \texttt{</instruction>}
    } \\
    \bottomrule
    \end{tblr}
    \label{tab:benchmark_prompt_validate}
\end{table*}

\end{document}